\documentclass[lettersize,journal]{IEEEtran}
\usepackage{amsmath,amsfonts,amssymb,amsthm}
\usepackage{algorithmic}
\usepackage{array}
\usepackage[caption=false,font=normalsize,labelfont=sf,textfont=sf]{subfig}
\usepackage{textcomp}
\usepackage{stfloats}
\usepackage{url}
\usepackage{verbatim}
\usepackage{graphicx}
\usepackage{tabularx,booktabs}
\usepackage{comment}
\def\BibTeX{{\rm B\kern-.05em{\sc i\kern-.025em b}\kern-.08em
    T\kern-.1667em\lower.7ex\hbox{E}\kern-.125emX}}
\usepackage{balance}
\usepackage{adjustbox}
\usepackage{tikz}
\usepackage{pgfplots}
\pgfplotsset{compat=1.17}
\usepgfplotslibrary{fillbetween}

\pgfplotsset{
    every axis/.append style={
        tick label style={font=\normalsize},
        label style={font=\normalsize},
        legend style={font=\normalsize},
        title style={font=\normalsize}
    }
}
\usepackage{pgfplotstable}
\usetikzlibrary{pgfplots.groupplots}
\usetikzlibrary{pgfplots.dateplot}
\usetikzlibrary{arrows.meta, positioning}

\DeclareMathOperator*{\argmax}{arg\,max}
\DeclareMathOperator*{\argmin}{arg\,min}
\allowdisplaybreaks

\newcommand {\cmatr}[2]{\left\{\begin{array}{#1}#2\end{array}\right.}

\newcommand{\vect}[1]{\ensuremath{\boldsymbol{\mathrm{#1}}}}
\newtheorem{theorem}{Theorem}
\newtheorem{Proposition}{Proposition}
\newtheorem{corollary}{Corollary}
\newtheorem{Lemma}{Lemma}
\newtheorem{Definition}{Definition}

\newtheorem{Assumption}{Assumption}
\usepackage[nolist]{acronym}
\begin{acronym}
    \acro{sdm}[SDM]{Sequential Decision Making}
   \acro{mdp}[MDP]{Markov Decision Process}
   \acro{rl}[RL]{Reinforcement Learning}
   \acro{dp}[DP]{Dynamic Programming}
   \acro{adp}[ADP]{Approximate Dynamic Programming}
   \acro{sp}[SP]{Stochastic Programming}
   \acro{lp}[LP]{Linear Programming}
   \acro{mpc}[MPC]{Model Predictive Control}
   \acro{empc}[EMPC]{Economic Model Predictive Control}
   \acro{td}[TD]{Temporal Difference}
   \acro{mbrl}[MBRL]{Model-based Reinforcement Learning}
   \acro{dt}[DT]{Digital Twin}
   \acro{ml}[ML]{Machine Learning}
   \acro{lstm}[LSTM]{Long Short-Term Memory}
   \acro{rnn}[RNN]{Recurrent Neural Network }
   \acro{dnn}[DNN]{Deep Neural Network }
   \acro{dnns}[DNNs]{Deep Neural Networks}
   \acro{gp}[GP]{Gaussian Processes}
   \acro{ai}[AI]{Artificial Intelligence}
   \acro{mle}[MLE]{Maximum Likelihood Estimation}
   \acro{lqr}[LQR]{Linear Quadratic Regulator}
   \acro{gpu}[GPU]{Graphics processing unit}
   \acro{tpu}[TPU]{Tensor processing unit}
   \acro{gans}[GANs]{Generative Adversarial Networks}
   \acro{cnns}[CNNs]{Convolutional Neural Networks}
   \acro{ls}[LS]{Least Squares}
   \acro{mse}[MSE]{Mean Squared Error}
    \acro{map}[MAP]{Maximum A Posteriori}
    \acro{gmm}[GMM]{Gaussian Mixture Model}
    \acro{hmm}[HMM]{Hidden Markov Model}
    \acro{elbo}[ELBO]{Evidence Lower Bound}
    \acro{vae}[VAE]{Variational Autoencoder}
    \acro{gan}[GAN]{Generative Adversarial Network}
    \acro{ode}[ODE]{Ordinary Differential Equations}
    \acro{ude}[UDE]{Universal Differential Equations} 
    \acro{ham}[HAM]{Hybrid Analysis and Modeling} 
    \acro{vi}[VI]{Variational Inference} 
    \acro{em}[EM]{Expectation-Maximization} 
    \acro{lstdq}[LSTDQ]{Least Square Temporal Difference Q-learning}
    \acro{mppi}[MPPI]{Model Predictive Path Integral} 
    \acro{cem}[CEM]{Cross Entropy Method} 
    \acro{llm}[LLM]{Large Language Models} 
\end{acronym}

\usepackage{float}
\usepackage{color}
\usepackage{multicol}
\usepackage{multirow} 

\begin{document}
\title{All AI Models are Wrong, but Some are Optimal}
\author{Akhil S Anand, Shambhuraj Sawant, Dirk Reinhardt, Sebastien Gros 
\thanks{All the authors are with the Dept. of Engineering Cybernetics at Norwegian University of Science and Technology (NTNU), Trondheim, Norway.  {Contact: \tt\small akhil.s.anand@ntnu.no}}}
\maketitle
\begin{abstract}
  AI models that predict the future behavior of a system (a.k.a. predictive AI models) are central to intelligent decision-making. However, decision-making using predictive AI models often results in suboptimal performance. This is primarily because AI models are typically constructed to best fit the data, and hence to predict the most likely future rather than to enable high-performance decision-making. The hope that such prediction enables high-performance decisions is neither guaranteed in theory nor established in practice. In fact, there is increasing empirical evidence that predictive models must be tailored to decision-making objectives for performance. In this paper, we establish formal (necessary and sufficient) conditions that a predictive model (AI-based or not) must satisfy for a decision-making policy established using that model to be optimal. We then discuss their implications for building predictive AI models for sequential decision-making. 
\end{abstract}

\section{Introduction}


With advances in \ac{ai}, computational power, and big data, it has become increasingly possible to build data-driven, predictive AI models that can predict the future behavior of complex physical systems and processes in response to actions we plan to apply in reality \cite{emmert2020introductory}.  These predictive models play a central role in intelligent systems by facilitating autonomous decision-making \cite{rasheed2020digital}. Despite their potential, predictive AI models are not yet widely adopted for sequential decision-making, in part due to the disappointing performance of the resulting decisions. This performance gap is due to the fact that most real systems are inherently stochastic and, even with abundant data, the predictions of the AI model are always an approximation of the real system's future behavior \cite{botin2022digital}.

Predictive \ac{ai} models for decision-making, whether probabilistic or deterministic nature, are typically constructed to maximize prediction accuracy by fitting the model to the observed behavior of the physical system from data \cite{rasheed2020digital, agrell2023optimal}. When used for decision-making, the hope is that the best-fitted predictive model, according to some chosen loss function, will enable decisions that are close to optimal. However, when dealing with stochasticity, this assumption is not well-supported in theory and is often challenged in practice \cite{gros2019data, piga2019performance}. This limitation arises because the construction of the predictive models is generally agnostic to the decision-making objectives, and therefore has no direct relationship to the performance measure of the resulting decision-making scheme.

In practice, this is evident in \ac{mbrl}, where even complex \acp{dnn} predictive models optimized for accurate predictions often fail to adequately represent the real system, becoming a bottleneck for decision-making performance \cite{moerland2023model}. In fact, there is growing empirical evidence that embedding decision-making objectives in predictive models can significantly enhance decision-making performance \cite{pmlr-v162-hansen22a, farahmand2017value, schrittwieser2020mastering, gros2019data, kordabad2023reinforcement}. Additionally, it has been observed that predictive models can trade prediction accuracy for improved decision-making performance \cite{larsen2024variational, menges2024digital}. A parallel can be drawn with training \ac{llm}, where \ac{rl} with human feedback is applied post-training to refine performance beyond data-fitting \cite{retzlaff2024human}. Despite the empirical evidence, the formal relationship between predictive \ac{ai} models and the performance of the decisions one can draw from these models remains unknown. This paper aims at filling that gap.

Building on prior works in the context of \ac{mpc} \cite{anand2024data,anand2024optimal}, in this paper, we present a formal framework to construct predictive AI models tailored to decision-making performance instead of prediction accuracy, referred to as ``decision-oriented'' predictive models. The main contributions of this paper are:
\begin{itemize}
  \item Establish the necessary and sufficient conditions for decision-oriented predictive AI models and discuss their implications for constructing such models.
  \item Prove that a predictive AI model best fitting the data is not necessarily the AI model that enables the best decision-making performance. 
  \item Show how deterministic predictive models can enable optimal decision-making for stochastic systems.
  \item Establish the class of decision-making problems where AI models built to fit the data enable optimal decisions and in what sense.
\end{itemize}

The rest of the paper is structured as follows: Section \ref{sec:stochastic_sdm} introduces the necessary concepts within sequential decision-making for stochastic systems. Section \ref{sec:dt for sdm} formulates the predictive model-based decision-making framework and discusses different classes of predictive models and different methods to estimate them. Section \ref{sec:theorem} introduces the central concepts in the paper and specifies the necessary and sufficient conditions on predictive models to achieve optimal decision-making. Section \ref{sec:examples} provides two sets of simulation studies to demonstrate the concept of constructing decision-oriented predictive models. Section \ref{sec:discussion} provides a discussion of our findings. Section \ref{sec:conclusion} presents the conclusions.

\begin{figure}[t!]
  \centering
  \includegraphics[width= 1\columnwidth]{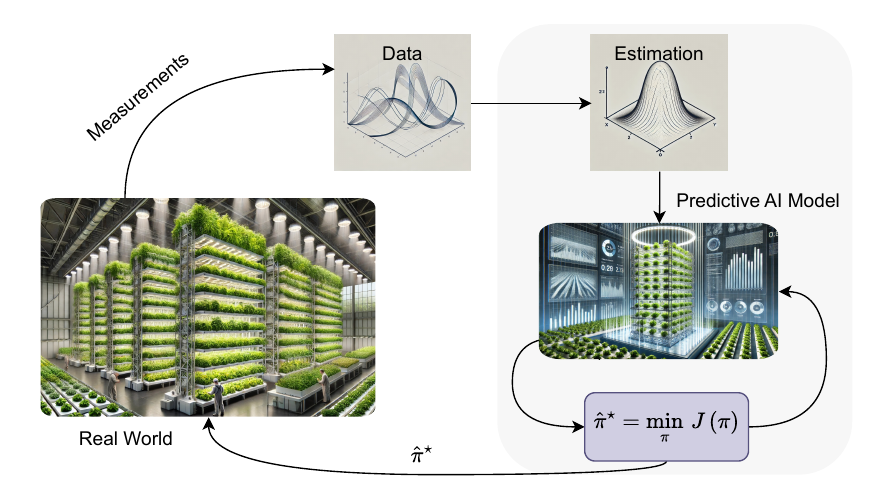}
  \caption{Model-based decision-making concept presented with the example of an indoor vertical farm. The standard model-based decision-making framework involves estimating a predictive model from data to derive optimal decisions for the real system. Then this predictive model is used in silico to generate the best possible decisions for a given decision-making objective. The model-based decisions are then implemented on the real system.}
  \label{fig:dm_framework}
\end{figure}

\section{Background on sequential decision-making} \label{sec:stochastic_sdm}
In sequential decision-making, an agent aims to maximize expected rational utility over time by making decisions based on the latest available information, while accounting for both short- and long-term consequences \cite{littman1996algorithms}. For stochastic systems, decisions must also account for future uncertainty and the fact that more information will be available for future decisions. We use the terms decision-making and sequential decision-making interchangeably throughout this paper. 

\subsection{Sequential Decision-Making Problem as MDP}
The \ac{mdp} framework provides a solid foundation to formulate and solve sequential decision-making problems in the context of stochastic systems with the Markovian property \cite{puterman2014markov}. \ac{mdp}s consider dynamic systems with underlying  states $\vect s\in \mathbb S$ and actions $\vect a \in \mathbb A$, with respective sets $\mathbb S$ and $\mathbb A$, and the associated stochastic state transition:
\begin{align}
\label{eq:StateTransition}
\vect s_+ \sim \rho \left(\,.\,|\,\vect s, \vect a\,\right)\,,
\end{align}
defining how a state-action pair $\vect s, \vect a$ yields a new state $\vect s_+$. 
In \eqref{eq:StateTransition}, $\rho$ can be of different nature, depending on the set $\mathbb S$. 
For $\mathbb S$ being a discrete countable set, $\rho$ is a conditional probability. 
For $\mathbb S\subseteq \mathbb R^n$, where $n$ is the state dimension, $\rho$ can be a conditional probability density or a conditional probability measure.

Solving an \ac{mdp} consists in finding a (possibly stochastic) policy $\pi\left(\vect a|\vect s\right)$ such that the stochastic closed-loop trajectories maximize the expected value of a given long-term reward of actions specified by $r\,:\, \mathbb S \times  \mathbb A\,\mapsto \, \mathbb R$. The most common criterion is maximizing the sum of discounted rewards:
\begin{align}
\label{eq:MDPCost}
J (\pi) = \mathbb E\left[\left.\sum_{k=0}^\infty \gamma^k r\left(\vect s_k,\vect a_k\right)\,  \right |\, \vect a_k\sim \pi(.|\vect s_k)\right]\,,
\end{align}
for a discount factor $\gamma \in (0,1)$. The $\gamma$ can be viewed as a rudimentary probabilistic model of the system lifetime, meaningful to most applications. The expectation in \eqref{eq:MDPCost} is taken over the distribution of states \( \vect s_k \) and actions \( \vect a_k \) in the Markov chain induced by $\pi (\vect a | \vect s)$. The complete \ac{mdp} can  be represented as a tuple ${({\mathbb{S}},\,{\mathbb{A}},\,{r},\, {\rho},\gamma)}$. The solution to MDP provides an optimal policy $\pi^\star$ from the set $\Pi$ of all admissible policies by maximizing \eqref{eq:MDPCost}, defined as: 
\begin{align}
\label{eq:OptPolicy}
\pi^\star = \argmax_{\pi\in\Pi}\, J\left(\pi\right).
\end{align}
The process of deriving the policy is termed as ``policy optimization''. Note that classical \acp{mdp} always admits a deterministic policy (\({\pi}: \mathbb{S} \to \mathbb{A}\)) as an optimal solution. Only in fairly exotic problem settings, such as certain partially observable or constrained \acp{mdp} admit a stochastic policy as the optimal solution \cite{shen2024flipping}.

\subsection{Solving the MDP}\label{sec:MDP_solution}
The solution of an \ac{mdp} is described through the underlying Bellman equations \cite{bellman1957dynamic}:
\begin{subequations}
\label{eq:Bellman}
\begin{align}
Q^\star\left(\vect s,\vect a\right) &= r\left(\vect s,\vect a\right) + \gamma \,\mathbb E_{\rho}\left[ V^\star\left(\vect s_+\right)\,|\, \vect s,\vect a \right] \,,\\
V^\star\left(\vect s\right) &= \max_{\vect a}\,\, Q^\star\left(\vect s,\vect a\right) \label{eq:Vstar:Bell} \,, \\
\pi^\star\left(\vect s\right) &= \mathrm{arg} \max_{\vect a}\,\, Q^\star\left(\vect s,\vect a\right) \,, \label{eq:Pistar:Bell}
\end{align}
\end{subequations}
where $V^\star$ and $Q^\star$ are the optimal value and action-value functions, respectively. $\mathbb{E}_{{\rho}}\left[\cdot \mid \vect{s}, \vect{a}\right]$ represents the expectation over the distribution \eqref{eq:StateTransition}. It is useful to observe here that $Q^\star$ is arguably the most informative object in \eqref{eq:Bellman} since it implicitly defines the other ones. 

The Bellman equations can be solved iteratively using policy optimization methods such as \ac{dp}\cite{puterman2014markov}, value iteration, or policy iteration. But these approaches are computationally challenging, due to the "curse of dimensionality" \cite{powell2007approximate}, hence in practice, most \acp{mdp} are solved approximately \cite{powell2007approximate}. Standard algorithms for policy optimization are: \ac{adp} which includes \ac{rl}), Monte Carlo Tree Search,  \ac{mpc}, \ac{mppi}, \ac{cem} etc. Notably, a special case of \(\rho\) in \eqref{eq:StateTransition} admits a linear and closed-form solution to \(\pi^\star\): linear dynamics with additive process noise (of any kind) and a quadratic reward function.

\subsection{Economic vs Tracking Problems}
Based on the nature of the optimization objective, decision-making problems can be classified into two categories that will prove important here: (i) tracking problems and  (ii) economic problems. The distinction between economic and tracking problems is well identified in the \ac{mpc} community \cite{zanon2016tracking, rawlings2012fundamentals}. Tracking problems are a common class of decision-making problems whose general objective is to track a reference state or trajectory by steering the system to an optimal steady-state or optimal trajectory defined by a feasible input-state reference.  It is typically formulated by defining a convex, preferably quadratic objective function that penalizes deviations of the states from the reference. Tracking problems constitute a major class of problems in robotics, process control, etc. In contrast, a decision-making problem is classified as \textit{economic} if the objective is to optimize an economic metric, which is a direct or indirect measure of profit maximization, cost or energy minimization, or maximizing return on investment, etc.  Economic problems are common in, e.g. finance,  energy management, logistics, manufacturing processes, etc. 

In economic problems, the decision-making objective typically represents the physical or the true quantity being optimized, such as energy, money, or resources. In contrast, tracking objectives usually represent a proxy of the true objective and guide the system toward an advantageous steady state. An economic objective is not necessarily lower-bounded, whereas tracking objectives are. Furthermore, in tracking problems, stability and constraint satisfaction are often more critical than optimality, whereas optimality is the most important criterion in economic problems. 


\section{Predictive Models for Decision-Making}\label{sec:dt for sdm}
A predictive model $\hat \rho \,:\, \hat {\mathbb S} \times  { \mathbb A}\,\mapsto \,  {\mathbb S}$, approximates the state transition probability of the real system \eqref{eq:StateTransition} as: 
\begin{equation}\label{eq:DT_dist}
  \hat {\vect s}_+ \sim \hat \rho \left(\,.\,|\,\vect s, \vect a\,\right)\,.
\end{equation}
Where $\hat {\vect s}_+ $ represents the states of the predictive model. Note that the states of the real system ($\vect s$) are also part of $\hat {\mathbb S}$ as the predictive model can be initialized with $\vect s$. We assume that the state of the system is known, but in cases where the state is not directly observable, we rely on the assumption that a sufficiently long input-output history is Markovian, allowing us to infer the necessary states. The predictive model \eqref{eq:DT_dist} is then used to support a model-based \ac{mdp}:
\begin{equation}\label{eq:MDP_DT}
  {({\mathbb{S}},\,{\mathbb{A}},\,{r},\, \hat{\rho},\gamma)}\,,
\end{equation}
and to form the corresponding optimal decision policy $\hat {\pi}_\star$, optimal action-value function $\hat{Q}^\star$, and optimal value function  $\hat{V}^\star$. The hope is that $\hat {\pi}_\star$ will perform well on the real \ac{mdp} associated with the real system.

The conventional view on this approach is that if the predictive model $\hat{\rho}$ captures the real system dynamics accurately, at least around the state distribution underlying the optimal decision policy, then decisions established from the model \ac{mdp} will perform well on real system \ac{mdp}. Accurate modeling is nearly impossible due to the inherent difficulty of approximating true densities, which constitute infinitely many moments. Therefore, simpler and partial statistics, such as expected values (1st moment), are far easier to predict and are commonly used in practice.  Additionally, predictive modeling is limited by the difficulties in, capturing uncertainties, noisy, and limited data. Uncertainties can arise fundamentally in two ways: (i) stochasticity of the real system seen as measurement noise (aleatoric uncertainty) and (ii) epistemic uncertainty which arises from the lack of data. Quantifying and separating the two types of uncertainties remain a major challenge in predictive modeling \cite{der2009aleatory}. Due to this inevitable mismatch between $\hat{\rho}$ and the real system \eqref{eq:StateTransition}, the solution of the model-based \ac{mdp} \eqref{eq:MDP_DT} may differ from that of the real \ac{mdp} \cite{moerland2023model}.

\subsection{Classes of Predictive Models}
Predictive models can be classified based on the nature of their underlying predictive mechanism: (i) whether their predictions are probabilistic or deterministic, and (ii) whether they predict a single-step or multiple steps into the future.

\subsubsection{Probabilistic vs deterministic predictive models}
Probabilistic models \(\hat{\rho} \,(\hat{\vect s}_+ | \vect s, \vect a )\) represent the conditional distribution of future states, allowing the quantification of uncertainty. They can either model the full state distribution (analytical models) or provide samples (generative models). When used for decision-making analytical models often involve computationally intractable integrations over high-dimensional distributions, whereas, sampling-based models simplify decision-making by drawing samples to evaluate expectations. For stochastic environments, computational complexity and uncertainties make probabilistic modeling challenging unless the distribution is of a tractable form, such as Gaussian \cite{chua2018deep, deisenroth2011pilco}. In the case of deterministic models, \( \hat{\rho}\) reduces to a Dirac measure, denoted as \(\vect f (\vect s, \vect a)\), effectively collapsing the probabilistic distribution to a single-point prediction. These models are simpler to construct using regression techniques and easier to handle in decision-making as they enable iterative planning in a computationally efficient manner by bypassing the need to sample or integrate over probability distributions.

\subsubsection{Single-step vs Multi-step Predictive Models}
Predictive models can be classified as \textit{single-step} or \textit{multi-step} based on their prediction horizon. single-step models, such as state-space models, predict the future in a one-step ahead manner, \(\hat{\rho} \,(\vect s_{k+1} | \vect s_k, \vect a_k )\). Such single-step models can be used in an autoregressive fashion for multi-step predictions (e.g., \ac{rnn}, \ac{lstm}). In contrast, multi-step models, directly predict multiple future states, \(\hat{\rho}\, (\vect s_{k+1:N} | \vect s_k, \vect a_{k:N} )\), avoiding autoregression and the accumulation of single-step errors \cite{taieb2012review}.

\subsection{Parameter Estimation Methods}\label{sec:statistics}
Parameter estimation approaches for predictive models range from classical system identification \cite{ljung1999system} to modern machine learning techniques \cite{hastie2009elements, goodfellow2016deep}. Deterministic models commonly use expected-value estimation, which minimizes the expected value of a loss function for e.g. \ac{mse} loss, and estimates the conditional mean:
\begin{align}
  \label{eq:E:Fitting}
  {\vect f}_{\vect \theta}\left(\vect s,\vect a\right) = \mathbb E_{\rho}\left[\rho\left(\vect s_+\,|\,\vect s,\vect a \right)\right]\,.
\end{align}
This approach includes linear regression and variants like Lasso and Ridge regression \cite{yu2020prediction} which provides a point estimate that ideally matches the expected value of the real distribution. Probabilistic models usually employ \ac{mle} or Bayesian estimation approaches.  \ac{mle} maximizes the likelihood of observed data:

\begin{align}
  \label{eq:MLE:Fitting}
    \hat{\rho}_{\vect \theta}\left(\hat{\vect s}_+ | \vect s,\vect a\right) = \argmin_{\hat{\rho}} \sum_{i=1}^N -\log \rho\left(\vect s_+^{(i)}\,|\,\vect s^{(i)}, \vect a^{(i)}\right)\,.
  \end{align}

\begin{figure}[t]
  \centering
  \resizebox{0.8\columnwidth}{!}{%
    \begin{tikzpicture}
    \begin{axis}[
        width=\columnwidth,
        height=6cm,
        xlabel={${\vect s}_+$},
        ylabel={$\rho(\vect{s}_+ \mid \vect{s}, \vect{a})$},
        ymin=0,
        ymax = 0.22,
        xmin=0,
        xmax=10,
        legend pos=north west,
        legend style={draw=none, font=\small},
        yticklabels={},
        xticklabels={}, 
        xtick=\empty, 
        axis lines=left,
        grid=none
    ]
    
    \addplot[black, thick] table [x=x, y=density, col sep=comma] {Figures/E_vs_MLE/right_skewed_bimodal_density.csv};

    \pgfplotstableread[col sep=comma]{Figures/E_vs_MLE/mean_mle_points.csv}\datatable
    \pgfplotstablegetelem{0}{x}\of{\datatable}\edef\meanX{\pgfplotsretval}
    \pgfplotstablegetelem{0}{density}\of{\datatable}\edef\meanY{\pgfplotsretval}
    \node[anchor=west, text=black, yshift=0cm] at (axis cs:\meanX, \meanY) {$\mathbb E_{\rho}\left[\rho\left({\vect s}_+\,|\,\vect s,\vect a \right)\right]$};

    \pgfplotstablegetelem{1}{x}\of{\datatable}\edef\mleX{\pgfplotsretval}
    \pgfplotstablegetelem{1}{density}\of{\datatable}\edef\mleY{\pgfplotsretval}
    \node[anchor=west, text=black,  xshift=-1.5cm, yshift=0.4cm] at (axis cs:\mleX, \mleY) {$\argmax_{{\vect s}_+} \rho\left({\vect s}_+\,|\,\vect s,\vect a \right)$};

    \fill[black] (axis cs:\meanX, \meanY) circle (2pt);
    \fill[black] (axis cs:\mleX, \mleY) circle (2pt);

    \end{axis}
\end{tikzpicture}
  }
  \caption{Expected-value and \ac{mle} of a probability density function $\rho$.}
  \label{fig:StatModel}
\end{figure}
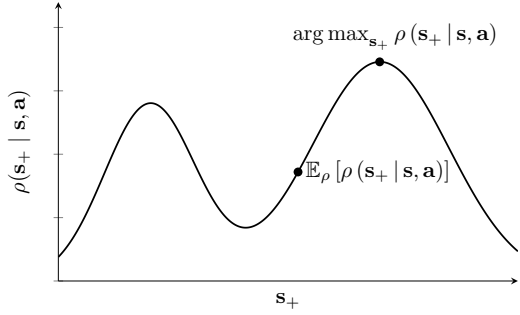

The fundamental difference between expected-value and \ac{mle} approaches is illustrated in  Fig.~\ref{fig:StatModel}. \ac{map} estimation approach offers an alternate by incorporating prior knowledge on the predictive model parameter and can be viewed as a regularized form of \ac{mle} with the prior term acting as the regularizer. Bayesian estimation extends this further by defining a posterior distribution over the model parameters based on Baye's rule.

There are countless model architectures and a myriad of methods to estimate their parameters,  but they fundamentally approximate a few key statistical properties of the data. Indeed, the statistics these approaches estimate can often be understood through their corresponding loss functions. Approaches using \ac{mse} loss are trying to estimate the expected mean, loss functions such as likelihood, KL divergence, and the cross-entropy loss, are estimating the likelihood. In contrast, Bayesian estimation is different; it is not straightforward to link it to any specific prediction statistics. However, it is reasonable to suggest that the Bayesian approach primarily aims to estimate the likelihood statistics in practice, while representing it as probability densities in the parameter space. Therefore, likelihood and expectation are arguably the two key statistics a predictive model is trying to build in most cases.  Additionally,  there are other less important approaches, such as Quantile Regression and adversarial learning \cite{calvo2023collaborative, zhao2020bridging}, which do not yield clear statistical measures and are therefore not relevant to the proposed framework.

\section{Optimal decision-making with decision-oriented predictive models}\label{sec:optimality}

In this section, we present the necessary and sufficient conditions for predictive \ac{ai} models to enable optimal decision-making. The theoretical results outlined here closely parallel the findings in the context of \ac{mpc} \cite{anand2024optimal}. We will start by clarifying the conditions on the equivalency between the solution to model-based \ac{mdp} and the real \ac{mdp} within the Bellman optimality framework \eqref{eq:Bellman}. The optimal action-value function associated with model-based \ac{mdp} \eqref{eq:MDP_DT} with an optimal policy ${\hat{\pi}}^\star$, can be expressed in terms of the reward function $r$ and the optimal value function $\hat{V}^\star$ as: 
\begin{equation}
    \hat{Q}^\star (\vect s, \vect a) = r (\vect s, \vect a) + \gamma \mathbb{E}_{\hat{\rho}}\left[\hat{V}^{\star}\left(\vect{\hat{s}}_{+}\right) \mid \vect{s}, \vect{a}\right] \,.
\end{equation}
$\mathbb{E}_{\hat{\rho}}\left[\cdot \mid \vect{s}, \vect{a}\right]$ represents the expectation over the model distribution \eqref{eq:DT_dist}. We now introduce the following technical assumption which is central to the discussion on optimality.
\begin{Assumption}
\label{assum:technical_assumption}
The set
\begin{equation} \label{eq:bounded_V}
\Omega : \Bigg\{ \vect s \in \mathcal{S} \, \Bigg |\,\left|\mathbb{E}_{\hat{\rho}}\left[\hat{V}^{\star}\left(\vect{\hat{s}}_k^{\pi^{\star}}\right)\right]\right|<\infty,  \quad \forall\, k < N \Bigg\}
\end{equation}
is assumed to be non-empty for stochastic trajectories of the model \((\vect{\hat{s}}_0^{\pi^{\star}}, \dots, \vect{\hat{s}}_N^{\pi^{\star}})\) under the optimal policy \(\pi^\star\).
\end{Assumption}

This assumption requires the existence of a non-empty set such that the optimal value function $\hat{V}^\star$ of the predicted optimal trajectories \((\vect{\hat{s}}_0^{\pi^{\star}}, \dots, \vect{\hat{s}}_N^{\pi^{\star}})\) on the system model is finite for all initial states starting from this set. The assumption \eqref{eq:bounded_V} requires that the predictive model trajectories under the optimal policy ${\pi}^\star$ are contained within the set $\mathcal{S}$ where the value function ${V}^\star$ is bounded with a unitary probability. This can be interpreted as a form of stability condition on $\hat{\rho}$ under the optimal trajectory \((\vect{\hat{s}}_0^{\pi^{\star}}, \dots, \vect{\hat{s}}_N^{\pi^{\star}})\).

In $\mathcal{S}$ for any action $\vect a$ such that $\left|\mathbb{E}_{\hat{\rho}}\left[\hat{V}_{\star}\left(\hat{\vect s}_{+}\right) \mid \vect s, \vect a\right]\right|<\infty$, the optimality condition can be represented in terms of the optimal action-value function $Q^\star$ as follows. If
\begin{align}
\label{eq:DT:PerfectModel}
\hat Q^\star\left(\vect s,\vect a\right)=Q^\star\left(\vect s,\vect a\right),\quad \forall\,\vect s, \vect a \,, 
\end{align}
then the model-based \ac{mdp} \eqref{eq:MDP_DT} can provide ${\pi}^\star$, $V^\star$, and $Q^\star$ (of the real \ac{mdp}). In this sense, the model-based \ac{mdp} fully represents the real \ac{mdp}. However, we are primarily interested in the optimality of the resulting policy. Therefore, condition \eqref{eq:DT:PerfectModel} can be relaxed such that only policy $\hat{\pi}^\star$ needs to match the $\pi^\star$, not necessarily the $Q^\star$ and the $V^\star$. Therefore, $\hat{Q}^\star$ needs to match $Q^{\star}$ only in the sense that:
\begin{align}
\label{eq:DT:PerfectModel:argmin}
\mathrm{arg}\max_{\vect a}\,\hat{Q}^\star\left(\vect s,\vect a\right)=\mathrm{arg}\max_{\vect a}\,Q^\star\left(\vect s,\vect a\right),\quad \forall\,\vect s \,.
\end{align}

Note that \eqref{eq:DT:PerfectModel} implies \eqref{eq:DT:PerfectModel:argmin}, but the converse is not true, making \eqref{eq:DT:PerfectModel:argmin} less restrictive than \eqref{eq:DT:PerfectModel}. However, the less restrictive condition \eqref{eq:DT:PerfectModel:argmin}  is insufficient to ensure that the solution of the model-based \ac{mdp} fully satisfies the solution of the real \ac{mdp} in the sense of $V^\star$ and $Q^\star$.  An additional case of interest is, if \eqref{eq:DT:PerfectModel} is relaxed to be valid up to a constant, i.e., 
\begin{align}
\label{eq:DT:PerfectModel:PlusConstant}
\hat Q^\star\left(\vect s,\vect a\right) + Q_0  =Q^\star\left(\vect s,\vect a\right),\quad \forall\,\vect s,\vect a\,,
\end{align}
for some constant $Q_0\in\mathbb R$. If an model-based \ac{mdp} satisfies condition \eqref{eq:DT:PerfectModel:PlusConstant}, then it can deliver the optimal policy and the optimal value functions up to a constant. This is therefore a sufficient condition for the predictive model optimality. Note that \eqref{eq:DT:PerfectModel:PlusConstant} can be converted to condition \eqref{eq:DT:PerfectModel} simply by adding a constant to the costs in \eqref{eq:MDPCost}. 

Now we define the notion of predictive model optimality and decision-oriented predictive models:

\begin{Definition}
  A predictive model is ``optimal'' (or closed-loop optimal) for a decision-making task if the decisions derived from the corresponding model-based decision-making scheme satisfy \eqref{eq:DT:PerfectModel:argmin}, ensuring \(\hat{{\pi}}^\star\) = \({\pi}^\star\). 
\end{Definition}

Conditions  \eqref{eq:DT:PerfectModel} - \eqref{eq:DT:PerfectModel:PlusConstant} are trivial to represent the optimality of a model-based \ac{mdp}. However, they are central to establishing the necessary and sufficient conditions such that the predictive \ac{ai} model is optimal in the following section. Note that, the sufficient conditions, \eqref{eq:DT:PerfectModel} or \eqref{eq:DT:PerfectModel:PlusConstant} imply invoking $Q$-learning in the \ac{rl} context as it demands the resulting model-based decision-making scheme to approximate the $Q^\star$ in order to derive ${\pi}^\star$.  Whereas, the necessary condition, \eqref{eq:DT:PerfectModel:argmin} implies invoking policy gradient methods in the \ac{rl} context since it represents directly optimizing the model-based policy to match ${\pi}^\star$ without requiring to capture $Q^\star$.

\subsection{Conditions on Predictive Model Optimality} \label{sec:theorem}
Building on the conditions from the previous section now we will derive the the necessary and sufficient conditions on the predictive model such that it is optimal.  We start by considering the following modification to $\hat{V}^\star$ of the model-based \ac{mdp} with a choice of storage function $\lambda(\vect s)$, 
\begin{equation}\label{eq:modified_V}
    \hat{V}^\star(\vect s) \leftarrow \hat{V}^\star(\vect s) + \lambda(\vect s) \,,
\end{equation}
which results in the following modification to $\hat{Q}^\star$, 
\begin{equation}\label{eq:modified_Q}
  \hat{Q}^\star(\vect{s}, \vect{a}) \leftarrow \hat{Q}^\star(\vect{s}, \vect{a}) + \lambda(\vect{s}) \,.
\end{equation}

Note that the storage function $\lambda(\vect{s})$ can be chosen arbitrarily without affecting the optimal policy of the model-based \ac{mdp},
\begin{equation}
  \hat{\pi}^\star = \argmax_{\vect a} \hat{Q}^\star(\vect{s}, \vect{a})  = \argmax_{\vect a} \hat{Q}^\star(\vect{s}, \vect{a}) + \lambda(\vect s)\,.
\end{equation} 
Therefore, we will use this modified \ac{mdp} \eqref{eq:modified_Q} as a proxy for the model-based \ac{mdp} for optimality analysis. It is worth noting that \eqref{eq:DT:PerfectModel:PlusConstant} is a special case of \eqref{eq:modified_Q} where $\lambda(\vect s)$ a constant. We now define:
\begin{equation}\label{eq:Lambda}
    \Lambda (\vect s, \vect a) := \lambda(\vect s) - \gamma \mathbb{E}_{\rho} [\lambda(\hat{\vect{s}}_+) \mid \vect{s}, \vect{a}]\,.
\end{equation}
Given the Assumption~\ref{assum:technical_assumption}, the modified model-based \ac{mdp} \eqref{eq:modified_Q} satisfies:
\begin{equation}\label{eq:modified_bellman}
    \hat{Q}^\star(\vect{s}, \vect{a}) = r(\vect{s}, \vect{a}) + \Lambda(\vect{s}, \vect{a}) + \gamma \mathbb{E}_{\hat{\rho}} [\hat{V}^\star(\hat{\vect{s}}_+ \mid \vect{s}, \vect{a})] \,.
\end{equation}
In addition, we observe that the choice of $\lambda$ does not affect the optimal advantage function of the \ac{mdp} defined as: 
\begin{equation}
    \hat{A}^\star(\vect s) = \hat{Q}^\star(\vect s) - \hat{V}^\star(\vect s).
\end{equation}
The necessary optimality condition, if it exists, should match the optimal policy under true and model-based \acp{mdp} satisfying \eqref{eq:DT:PerfectModel:argmin}. For the sake of simplicity, let us assume that $Q^\star$ and $\hat{Q}^\star$ are bounded on a compact set and infinite outside of it. 

The advantage function associated with the real \ac{mdp} is given by:
\begin{align}\label{eq:advantage_functions}
    A^{\star}(\mathbf{s}, \mathbf{a}) =Q^{\star}(\mathbf{s}, \mathbf{a})-V^{\star}(\mathbf{s}) \, ,
\end{align}
by construction, it holds that:
\begin{equation}\label{eq:Advantage_equality}
    \max _{\mathbf{a}} \hat{A}(\mathbf{s}, \mathbf{a})=\max _{\mathbf{a}} A^{\star}(\mathbf{s}, \mathbf{a})=0 \quad \forall \vect s  \,.
\end{equation}
With these preliminaries established, the following Lemma provides the necessary and sufficient conditions for the optimality of the modified model-based \ac{mdp}.

\begin{Lemma}\label{lem:necessary_condition_A}
    There exist class $\mathcal{K}$ functions $\alpha, \beta$ such that
    \begin{equation}\label{eq:necesary_condition_A}
        \alpha (A^{\star}(\mathbf{s}, \mathbf{a})) \geq \hat{A}^\star(\mathbf{s}, \mathbf{a}) \geq \beta (A^{\star}(\mathbf{s}, \mathbf{a})), \,\,\forall\, \vect s,\vect a\,,
    \end{equation}
    is necessary and sufficient for
    \begin{equation}
        \underset{{\vect a}}{\arg \max }\, \hat{A}^\star(\mathbf{s}, \mathbf{a})=\underset{{\vect a}}{\arg \max }\, A^{\star}(\mathbf{s}, \mathbf{a}), \quad \forall \, \mathbf{s}\,,
    \end{equation}
    to hold.
\end{Lemma}
The proof of Lemma 1 is provided in the Appendix~\ref{apx:proof_necessary_lemma}. Considering that the modified \ac{mdp} is equivalent to the model-based \ac{mdp} for optimality analysis, we now present the necessary and sufficient conditions for predictive model optimality in the following Theorem.

\begin{theorem}
  \label{Th:necesary_condition}
  The necessary and sufficient condition on a predictive model \eqref{eq:DT_dist} such that the corresponding model-based \ac{mdp} yields an optimal solution to the \ac{mdp} defined by the real system dynamics \eqref{eq:StateTransition} (i.e., \(\hat{{\pi}}^\star\) = \({\pi}^\star\)), is stated as follows:
  
    \begin{equation}\label{eq:necessary_condition}
        \begin{aligned}
        \alpha (r & (\mathbf{s}, \mathbf{a})+ \gamma \mathbb{E}_{\rho}\left[V^{\star}\left(\mathbf{s}_{+}\right) \mid \mathbf{s}, \mathbf{a}\right]- V^{\star}(\mathbf{s})) \geq \\
        & r(\mathbf{s}, \mathbf{a})+ \Lambda(\mathbf{s}, \mathbf{a})+\gamma \mathbb{E}_{\hat{\rho}}\left[{V}^{\star}\left(\mathbf{\hat{s}}_{+}\right) \mid \mathbf{s}, \mathbf{a}\right]-{V}^\star(\mathbf{s}) \geq \\
        & \hspace{1cm} \beta (r(\mathbf{s}, \mathbf{a})+\gamma \mathbb{E}_{\rho}\left[ V^{\star}\left(\mathbf{s}_{+}\right) \mid \mathbf{s}, \mathbf{a}\right]- V^{\star}(\mathbf{s}))
        \end{aligned}
    \end{equation}
    for all $\vect s, \vect a$ with $\lambda\left(\vect s\right)$ chosen such that
    \begin{align}
    \label{eq:ValueFunctionMatching}
    \hat{V}^\star\left(\vect s \right) = V^\star\left(\vect s \right)\,,
    \end{align} 
    where $V^\star$ and $\hat{V}^\star$ are bounded on the set $\Omega$ as defined in Assumption \ref{assum:technical_assumption}. 
\end{theorem}

\begin{corollary}
  \label{Th:sufficient_condition}
  Considering that $\alpha$ and $\beta$ are both the identity function, and $\lambda$ as a constant, which results in a constant $\Lambda= \Delta$, simplifies the necessary condition \eqref{eq:necessary_condition} to the following sufficient condition:
    \begin{align}
    \label{eq:suficient_condition}
    \mathbb{E}_{\rho}\left[V^\star\left(\vect{s}_+\right)\,|\, \vect s,\vect a\, \right] - \mathbb{E}_{\hat{\rho}}\left[{V}^\star\left(\hat{\vect s}_+\right)\,|\, \vect s,\vect a\, \right]  = \Delta \,.
    \end{align}
\end{corollary}

The proof for Theorem \ref{Th:necesary_condition} is provided in Appendix \ref{apx:proof_theorem} and the proof of Corollary \ref{Th:sufficient_condition} is provided in Appendix \ref{apx:proof_sufficeint}. 

It is important to observe here that the necessary condition \eqref{eq:necessary_condition} is purely related to the properties of the \ac{mdp}, i.e. the stochastic dynamics \eqref{eq:StateTransition}, and the objective for decision-making \eqref{eq:MDPCost}. It does not depend on the decision-making scheme itself, as long as it provides the optimal model-based policy \(\hat{{\pi}}^\star\). The necessary condition \eqref{eq:necessary_condition} interlaces the optimal value function of the \ac{mdp} $V^\star$ and the predictive model $\hat{\rho}$ in a non-trivial way. Importantly, the necessary condition \eqref{eq:necessary_condition} and the sufficient condition \eqref{eq:suficient_condition} are \textit{unlike} the conventional criteria used for training a predictive model, such as those in \eqref{eq:E:Fitting} or \eqref{eq:MLE:Fitting}. This suggests that ``a decision-making scheme using predictive models constructed using such conventional approaches does not necessarily yield optimal decisions''.  These conditions form the foundations for constructing predictive models tailored to performance (decision-oriented predictive models) rather than prediction accuracy (best fit to the data). We now discuss a few key observations from Theorem~\ref{Th:necesary_condition}.

\subsubsection{Implications on the nature of the model}
The optimality conditions presented apply to both probabilistic and deterministic predictive models. One key consequence of Theorem~\ref{Th:necesary_condition} is stated in the following corollary.

\begin{corollary}\label{cor:deterministic}
    Under Assumption \ref{assum:technical_assumption}, there exists a deterministic predictive model of the form $\hat {\vect s}_+ = \vect f(\vect s, \vect a))$ that is optimal for a real \ac{mdp} with stochastic dynamics.
\end{corollary}

The proof of Corollary~\ref{cor:deterministic} is provided in Appendix~\ref{apx:proof:deterministic}. This observation speaks in favor of using deterministic predictive models for stochastic systems rather than probabilistic predictive models which are more difficult to handle in decision-making. Indeed, empirical evidence from the state-of-the-art \ac{mbrl} approach demonstrates that using a deterministic, decision-oriented predictive model outperforms all other model-based and model-free methods across diverse continuous control tasks and achieves performance on par with the state-of-the-art in image-based \ac{rl} tasks \cite{pmlr-v162-hansen22a}. 

This is related to the fact that the Theorem~\ref{Th:necesary_condition} accounts for aleatoric uncertainties through the expected value estimations $\mathbb{E}_{\rho}[.]$. However, in real-world decision-making problems, epistemic uncertainty can significantly impact value function estimates, leading to issues such as overestimation. To address this, epistemic uncertainties must be managed using probabilistic models and standard techniques such as Bayesian estimation, knowledge embedding, Monte Carlo (MC) dropout, bootstrapping, and other uncertainty quantification methods \cite{lockwood2022review}.

\subsubsection{Properties of the optimal predictive model}
It is important to note that the necessary and sufficient conditions stated in Theorem~\ref{Th:necesary_condition} and Corollary~\ref{Th:sufficient_condition} do not uniquely determine the parameters \(\vect{\theta}\) of the predictive model \(\hat{\rho}\). Specifically, for a given state-action pair \((\vect s, \vect a)\) and a chosen \(\alpha
\text{ and } \beta \), the necessary condition requires \(\hat{\rho}\,(\hat{\vect s}_+ | \vect s, \vect a)\) to belong to the level set: \(\left\{\vect{\hat s} \quad\mathrm{s.t.}\quad \eqref{eq:necessary_condition} \right\}\), and for sufficient condition to be in the level set: \(\left\{\vect{\hat s} \quad\mathrm{s.t.}\quad \eqref{eq:suficient_condition} \right\}\) for a chosen $\Delta$. This formulation introduces degrees of freedom in selecting \(\hat{\rho}\), which must be judiciously leveraged to achieve meaningful predictive models. A reasonable and practical criterion for choosing \(\hat{\rho}\,(\hat{\vect s}_+ | \vect s, \vect a)\) is that it lies within the support set of the real system dynamics \(\rho\,(\vect s_+ | \vect s, \vect a)\). In the following Proposition, we establish the existence of such a deterministic model under the support of real system dynamics.

\begin{Proposition}\label{prop:existance}
Suppose that $V^\star$ is continuous and that the true state transition \eqref{eq:StateTransition} has a fully connected support. Then there exists a model \(\vect f( \vect s, \vect a )\)  such that
\begin{align}\label{eq:Opt}
\mathbb E\left[V_\star\left(\vect s_+\right)|\vect s,\vect a\right] = V_\star\left(\vect f\left(\vect s,\vect a\right)\right)
\end{align}
and
\begin{align}\label{eq:Likelihood}
\rho\left(\vect f\left(\vect s,\vect a\right)|\vect s,\vect a\right) > 0
\end{align}
hold for all $\vect s,\vect a$.
\end{Proposition}
Proof of the Proposition \ref{prop:existance}is provided in Appendix~\ref{apx:proof:existance}. The condition \eqref{eq:Likelihood} specifies that the likelihood of an optimal model \(\vect f\) under the support of the real system dynamics is strictly positive. It is worth noting that when the real \ac{mdp} imposes hard constraints, the predictive model specified by \eqref{eq:Opt} captures these constraints through the unboundedness of $V\star$. While this observation is intuitive, it highlights that Theorem~\ref{Th:necesary_condition} aligns with the principles of robust optimization, effectively accounting for worst-case scenarios in model-based decision-making.

It is important to note that \eqref{eq:necessary_condition} defines the predictive model $\hat{\rho}$ for each $(\vect s,\vect a)$ pairs independently. This implies that the continuity of the resulting predictive model is not guaranteed from \eqref{eq:necessary_condition} alone. Constructing $\hat{\rho}$ based on \eqref{eq:necessary_condition} is highly correlated with the reward function $r$ indirectly through the constraint defined over the value function.  Therefore, the structure of the reward function $r(\vect s, \vect a)$ can influence the structure of the predictive model $\hat{\rho}$, thereby possibly resulting in a different structure of the underlying dynamics than data fitting approach \eqref{eq:StateTransition}. 

\subsubsection{Simulation-based vs. multi-step predictive models}

One key observation from the necessary and sufficient conditions \eqref{eq:necessary_condition} and \eqref{eq:suficient_condition} is that they are intrinsically tied to the single-step state transitions of the predictive model. Therefore, in the case of a single-step model used to form simulation-based predictions, it is straightforward to analyze the optimality of the resulting decision-making scheme under Theorem \ref{Th:necesary_condition}. However, these conditions do not readily extend to multi-step predictive models.  This stems from the fact that analyzing the optimality of a predictive model requires the value function of the model-based \ac{mdp} to satisfy the Bellman condition:
\begin{equation}\label{eq:consistency}
    \hat{V}^\star\left(\vect s\right) = \max_{\vect a}\,\,r\left(\vect s,\vect a\right) + \gamma \,\mathbb{E}_{\hat{\rho}}\left[\hat{V}^\star\left(\hat{\vect s}_+\right)\,|\, \vect s,\vect a\, \right]  \,.
\end{equation}

This implies that, for the sufficient and necessary condition to be applicable, the resulting model-based decision-making scheme needs to be cast as a time-invariant sequential decision-making process. However, a decision-making scheme using multi-step predictive models is not guaranteed to be cast as a sequential decision-making problem as discussed in \cite{anand2024data}. Nevertheless, the optimality condition for the model-based \ac{mdp} expressed in terms of the advantage functions \eqref{eq:necesary_condition_A} in Lemma~\ref{lem:necessary_condition_A} applies to both single-step and multi-step predictive models. However, it is not clear how to apply this condition \eqref{eq:necesary_condition_A} to construct predictive models.

\begin{figure*}[ht]
  \centering
  \includegraphics[width= 0.95\textwidth]{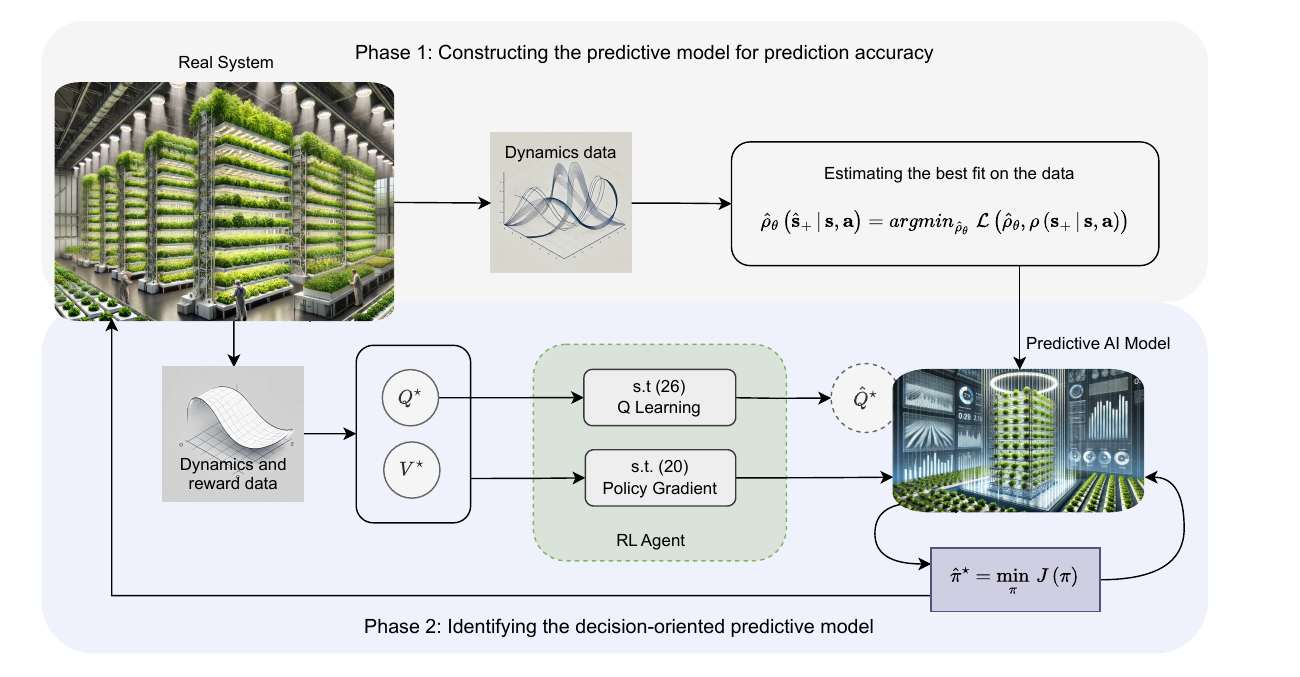}
  \caption{Constructing decision-oriented predictive models: The upper half (Phase 1) of the schematic illustrates the conventional approach to constructing predictive models, focusing on estimating the best fit to the real system's dynamics using the data and a choice of model estimation method. The lower half (Phase 2) represents the proposed approach to constructing decision-oriented predictive models using \ac{rl}. The key difference lies in the estimation approach, where the decision objective is embedded into the estimation process together with the support of the conventional data-fitting approach.}
  \label{fig:oo_framework}
\end{figure*}

\subsection{Constructing Decision-Oriented Predictive Models}\label{sec:idealmodel}
The necessary and sufficient conditions provide fundamental insight into how a predictive model ought to be tailored to enable optimal decision-making. We now discuss how these conditions can inform the development of an ``ideal'' approach for constructing decision-oriented predictive models.  Based on Proposition~\ref{prop:existance}, selecting a decision-oriented model based on the support of the real system transitions, while employing one of the model estimation methods discussed in Section~\ref{sec:statistics} offers a meaningful approach. For a well-specified decision-making problem, the parameters of the corresponding decision-oriented predictive model \(\hat{\rho}_{\vect \theta}(\hat{\vect s}_+ | \vect s, \vect a)\) that satisfies the sufficient or necessary condition, can be determined by solving the following constrained optimization problem.

\begin{equation}
\label{eq:Perf:SYSID}
\begin{aligned}
{\hat{\rho}_{\vect\theta}}\left(\hat{\vect s}_+\,|\,\vect s,\vect a\right) & = \argmin_{{\hat{\rho}}_{\vect{\theta}}}\,\, \mathcal{L}\left({\hat{\rho}}_{\vect{\theta}}, \rho\left(\vect s_+\,|\,\vect s,\vect a\right)\right) \\ 
& \mathrm{s.t.} \quad \quad \eqref{eq:necessary_condition} \quad \forall \, \vect s,\vect a\,,
\end{aligned}
\end{equation}

where \(\mathcal{L}(\cdot)\) represents a general loss function, which can be defined as \ac{mle} loss, Bayesian estimation loss, or expected-value losses such as Ridge or Lasso regression, as discussed in Section~\ref{sec:statistics}.

Solving the constrained optimization problem \eqref{eq:Perf:SYSID} in practice is challenging as it requires satisfying the sufficient or necessary condition explicitly. Indeed, condition \eqref{eq:necessary_condition} relies on the estimate of $V^\star$ and imposes non-trivial restrictions on the predictive model parameters $\theta$ which are potentially difficult to treat in practice. To address this, we propose leveraging \ac{rl} as a tool to estimate decision-oriented predictive models in practice. Instead of directly solving the constrained optimization problem, which can be computationally prohibitive,  \ac{rl} can be used to fine-tune a prior predictive model estimated for prediction accuracy (e.g., through~\eqref{eq:MLE:Fitting}) and iteratively improve its decision-making performance. In this approach, the model parameters are adjusted by interacting with the real system in a closed-loop manner for better performance. 

In the \ac{rl} context, both $Q$-learning and policy gradient-based \ac{rl} methods can be utilized. Policy gradient methods invoke the necessary condition \eqref{eq:necessary_condition}, while $Q$-learning methods invoke the sufficient condition \eqref{eq:suficient_condition}. In Q-learning, the objective is to construct a predictive model that minimizes the Bellman error between the $Q^\star$ and $\hat{Q}^\star$. The Q-learning approach optimizes the parameters of the predictive model \(\hat{\rho}_{\vect{\theta}}\) by minimizing \(\mathbb{E}_{\rho}\left[\left(Q^\star(\vect{s}, \vect{a}) - \hat{Q}^\star(\vect{s}, \vect{a})\right)^2\right]\). Note that the performance of a decision-oriented model derived through Q-learning is essentially limited by the richness of the function approximation for $Q^\star$.

Policy gradient methods directly optimize \(\hat{\rho}_{\vect{\theta}}\) to maximize the expected return \(J({\pi})\) \eqref{eq:MDPCost}, thereby deriving the optimal policy directly and aligns with the necessary condition. Policy gradient algorithms can take the form of deterministic policy gradient methods \cite{silver2014deterministic} or stochastic policy gradient methods \cite{sutton1999policy}. While stochastic policy gradients rely solely on the gradient of the model-based policy, deterministic policy gradients depend on the gradients of both the policy and the $Q$ function. Consequently, applying \ac{rl} in this manner requires the model-based decision-making scheme to be differentiable. This requirement is naturally satisfied in \ac{mpc} schemes, as demonstrated in Section~\ref{sec:smarthome}, due to their inherently differentiable structure. The performance of a decision-oriented model derived using policy gradient methods is limited by the richness of the policy parameterization, as it can only identify an optimal policy within the chosen parameterization class. Fig.~\ref{fig:oo_framework} represents the proposed approach to constructing decision-oriented predictive models using \ac{rl}. 

\ac{rl} is not necessarily the only solution to apply the theory in practice. We propose that embedding decision objectives into predictive models, in any form, can be helpful for constructing decision-oriented predictive models. One way to achieve this is by designing a suitable loss function for training the model, incorporating decision-oriented terms in addition to prediction accuracy terms.  Interestingly, many state-of-the-art \ac{mbrl} approaches aim to build decision-oriented predictive models that yield accurate value function estimates \cite{pmlr-v162-hansen22a, farahmand2017value, schrittwieser2020mastering} are effectively invoking the sufficient condition with $\Delta = 0$.  In \cite{abachi2020policy} a weighted maximum likelihood objective is employed, where the weights are selected to minimize the gap between the true policy gradient and the policy gradient in the model-induced \ac{mdp}, aligning with the necessary condition. In \cite{nikishin2022control} an end-to-end model learning approach that directly optimizes expected returns using implicit differentiation is adopted, mirroring the sufficient condition. This body of literature strongly supports our approach to constructing decision-oriented predictive models for high-performance decision-making.

\subsection{Local Optimality of Expected-Value Models}
\label{sec:LocalOpt}

In this section, we will discuss the conditions under which predictive models constructed using expected-value estimation \eqref{eq:E:Fitting}, referred to as expected-value models, can achieve local optimality in decision-making. This analysis closely parallels the discussion in the context of \ac{mpc}, as detailed in Lemma~1 of \cite{anand2024data}, which can be referred to for the technical details underpinning this section.

Expected-value models are particularly well-suited for decision-making problems where the system dynamics and objectives exhibit certain favorable properties. A prominent example is the \ac{lqr} problem, where the system dynamics are linear with additive Gaussian noise, and the cost function is quadratic, resulting in a quadratic and smooth value function. In such cases, the expected-value model approximating the true underlying system dynamics is optimal. 

While \ac{lqr} is a special case, the optimality of expected-value models can be generalized to \ac{mdp}s where the dynamics converge to a narrow steady-state region (its attraction set) and the value function remains smooth within this region. This corresponds to smooth and dissipative \acp{mdp} characterized by steady-state stochastic trajectories converging to a narrow set. There are two general classes of decision-making problems that fall into this category, where the first class is ``tracking problems'' with smooth dynamics and narrow steady-state distribution. For this class of problems, an expected-value model can be near-optimal at the steady-state distribution. This property explains the success of model-based decision-making schemes, such as \ac{mpc}, in solving complex tracking problems with expected-value models. Indeed, it can be argued that a not-so-perfect model can be good enough to achieve optimality if the underlying decision-making problem is a tracking problem \cite{gros2022economic}. 

The second class consists of ``economic'' \ac{mdp}s with smooth dynamics and smooth objective, achieving dissipativity \cite{gros2022economic} to a narrow attraction set. This highlights the role of dissipativity in achieving optimality for economic objective problems. It is worth noting that there exists an equivalence between tracking problems and dissipative economic problems. Specifically, an economic problem can be effectively reformulated as a tracking problem if the \ac{mdp} is dissipative; otherwise, it is not straightforward. For cases outside the classes of problems discussed, an expected-value model \eqref{eq:E:Fitting} does not necessarily enable the construction of an optimal policy, even locally. Such cases include, for example, \ac{mdp}s that do not exhibit dissipativity, \ac{mdp}s with non-smooth reward functions or non-smooth dynamics.

\section{Examples}\label{sec:examples}
In this section, we provide two sets of decision-making problems in simulation with varying complexity to demonstrate the implication of the Theory.  

\subsection{Example 1: Battery Energy Storage}\label{sec:BatExample}
In this simple example of a Battery energy storage system, we employ a model-based \ac{mdp} with an expected-value model to determine the optimal decisions for the real stochastic system. In this example, both the real \ac{mdp} and the model-based \ac{mdp} are solved through \ac{dp}. This example consists of two distinct cases: in case 1 we will demonstrate the suboptimality of a data-fitted model and in the second case we will additionally demonstrate the nature of the optimal model derived using the sufficient condition \eqref{eq:suficient_condition}.

Consider the following linear dynamics of a battery system with process noise,
\begin{align}
  \label{eq:Dynamics0}
  \vect s_+ = \vect s + \vect a + \vect w,\quad \vect w\sim \mathcal N\left(0,\sigma\right)\,,
\end{align}
where $\vect s$ is the relative stored energy, $\vect a$ is energy purchased or sold (e.g. from/to the electricity grid), and $\vect w$ is the imbalance in local energy consumption-production. Given the restrictions $\vect s\in [0,1]$ and  $\vect a\in [-0.25,0.25]$ and a discount factor $\gamma = 0.99$. The noise term $\vect w$ is Gaussian distributed with a standard deviation of $0.05$ and limited to the interval $\vect w\in[-0.05,0.05]$. The corresponding expected-value model of the system \eqref{eq:Dynamics0} is given by the linear model:
\begin{equation}\label{eq:eample1_nominal_model}
    \vect f(\vect s, \vect a) = \vect s+ \vect a\,.
\end{equation}

\subsubsection{Case 1}
Consider the reward,
\begin{align}
  \label{eq:PowerCost}
  r\left(\vect s,\vect a\right) = \cmatr{ccc}{- \vect a & \text{if} & \vect a \leq 0 \\ - 2\vect a&\text{if}&\vect a > 0}\,,
\end{align}
with the addition of high penalties for $\vect s\notin [0,1]$. The reward \eqref{eq:PowerCost} represents the different costs of buying ($\vect a > 0$) and selling  ($\vect a < 0$) energy. We observe that for all state $\vect s\in [0,1]$ there is a feasible action $\vect a\in [-0.25,0.25]$ that can keep $\vect s_+\in [0,1]$ regardless of the uncertainty $\vect w$. While using an expected-value model \eqref{eq:eample1_nominal_model} to solve the \ac{mdp},  we observe that even though the optimal value function $V^\star$ appears close to linear, but the resulting optimal model-based policy $\hat{\pi}^\star$ differs significantly from the optimal policy $\pi^\star$ of the real \ac{mdp} (see top row in Fig.~\ref{fig:BatVandPi}). That is because the optimal trajectories of the problem are not driven to a specific steady state in $[0,1]$, but rather cover a large part of the interval, and activate the lower bound of the state $\vect s=0$. Therefore an expected-value model is not optimal for this decision-making problem.

\subsubsection{Case 2} \label{sec:AbsValExample}
\begin{figure}[t]
    \centering
    \includegraphics[width=1\columnwidth]{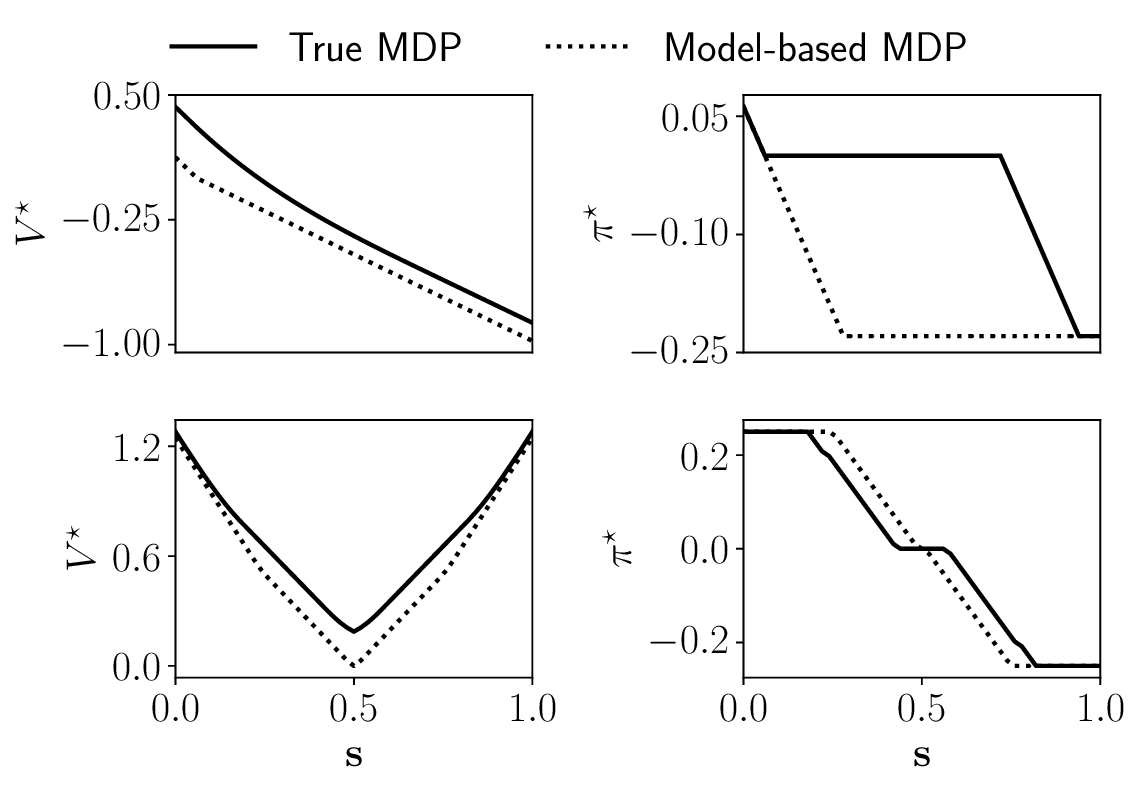} 
    \caption{(Top row) Illustration of the energy storage example~\ref{sec:BatExample}. The optimal value function (\(V^\star\)) and policy (\(\pi^\star\)) of the MDP are approximated by the \ac{mpc} as \(\hat{V}^\star\) and \({\hat{\pi}}^\star\), respectively. Although the value functions are nearly linear, the policies differ significantly due to the activation of the lower bound.  (Bottom row) A similar illustration of the non-smooth value function in example~\ref{sec:AbsValExample}.} 
    \label{fig:BatVandPi}
  \end{figure}

Consider the non-smooth reward:
\begin{align}
\label{eq:AbsCost} 
  r\left(\vect s,\vect a\right) = -\left|\vect s-\frac{1}{2}\right| -  \left|\vect a\right|
\end{align}
with the addition of high penalties for $\vect s\notin [0,1]$. In this example, $\vect w$ is sampled from a normal distribution with a standard deviation of $0.1$ and limited to the interval $\vect w\in[-0.25,0.25]$. We observe that for all states $\vect s\in [0,1]$ there is a feasible action $\vect a\in [-0.25,0.25]$ that can keep $\vect s_+\in [0,1]$ regardless of the uncertainty $\vect w$.  The bottom row in Fig.~\ref{fig:BatVandPi} shows the optimal value function $V^\star$, optimal policy $\pi^\star$, and the corresponding model-based values, ${\hat{V}}^\star$ and ${\hat{\pi}}^\star$. We observe that the optimal value function $V^\star$ is non-smooth at $\vect s=0.5$.  As a result, the model-based policy differs significantly from the optimal policy.

\begin{figure}[t]
  \centering
  \includegraphics[width=1\columnwidth]{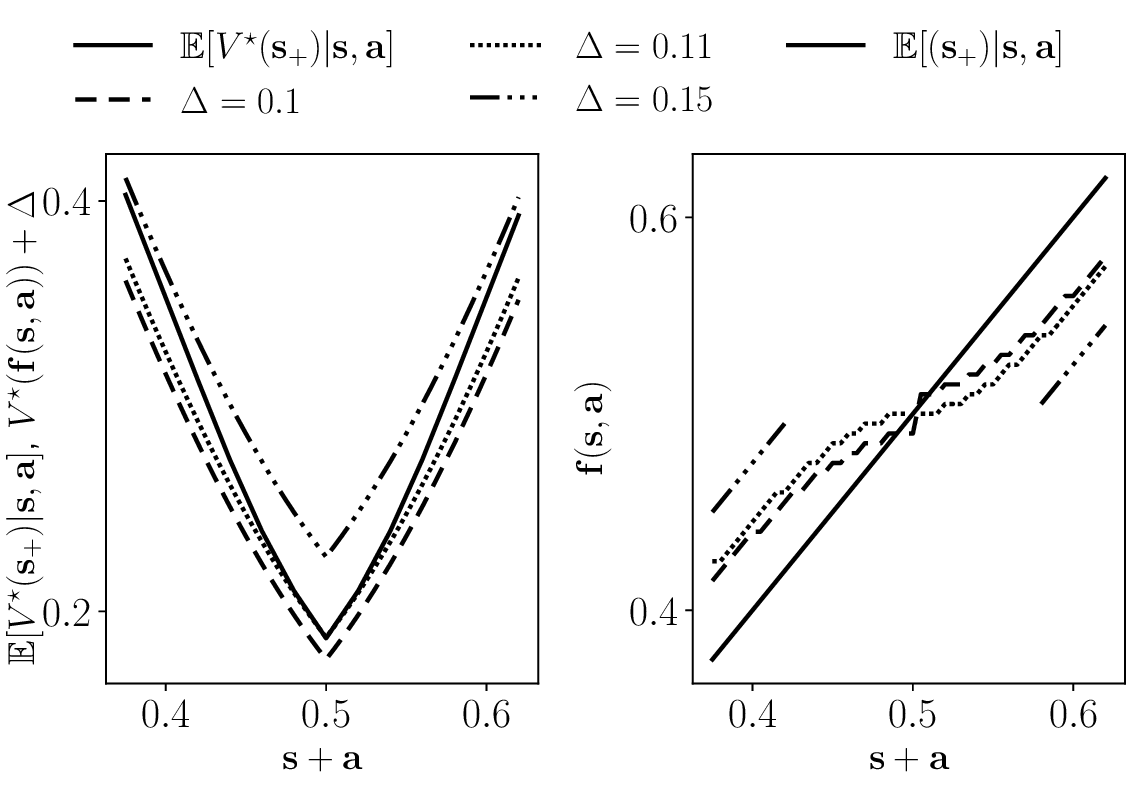} 
  \caption{Illustration of the models obtained from the sufficient condition \eqref{eq:suficient_condition} for different values of $\Delta$. The left plot shows the resulting value functions for the real \ac{mdp} and the model-based \ac{mdp} (\ac{mpc}). The right side plot shows the models in terms of $\vect s+\vect a$. The solid curve is the expected value of the real state transition, and the other curves are the optimal models for different values of $\Delta$. 
  }
      \label{fig:AbsDelta}
\end{figure}

For this example, we derived the optimal models using the sufficient condition \eqref{eq:suficient_condition} for three difference values of $\Delta$. The resulting models and their optimal value functions are shown in  Fig.~\ref{fig:AbsDelta}.  We observe that for $\Delta=0.15$, \eqref{eq:Perf:SYSID} produces a $\vect f$ that is not defined everywhere, despite the real system dynamics being fully defined. For  $\Delta= 0.1$, \eqref{eq:Perf:SYSID} yields a $\vect f$ that is not continuous everywhere, contrary to the real system dynamics being continuous everywhere. However, for a specific value of $\Delta=0.11$, the $\vect f$ from \eqref{eq:Perf:SYSID}  exists everywhere and is continuous, as seen in the green curve in Fig.~\ref{fig:AbsDelta}. However, this $\vect f$ is nonlinear even though the real system dynamics are linear in expected value.

\subsection{Example 2: Smart Home Heat Pump Control} \label{sec:smarthome}
Having demonstrated the theory in a simpler example in Section \ref{sec:BatExample}, this section presents a more practical demonstration of the theory through a simulation example of a smart home energy management problem. The major energy requirements for houses in colder or warmer climate regions are for heating and cooling, respectively. We consider a typical house located in central Trondheim (Norway) equipped with air-to-air heat pumps. A detailed account of the experimental house is available in the book chapter \cite{reinhardt2024data}. We utilize an \ac{mpc} as the decision-making scheme in this example.

\subsubsection{Smart home system}
Based on the experimental data from the house a data-driven simulation model of the house is constructed as an ARX model through subspace identification. The identified input-output model ARX model of smart house is represented as:

\begin{subequations} \label{eq:house}
\begin{align}
\vect{y}_{t+1} &= \vect f \begin{bmatrix} \vect{u}_p \\ \vect{y}_p \end{bmatrix} + \epsilon \sim \mathcal{N}(\mu, \sigma^2) \,, \label{eq:a} \\
\vect{u}_p &:= \vect{U}_{t-N_p+1, N_p, 1} = \begin{bmatrix} 
\vect{u}_{t-N_p+1} \\ 
\vect{u}_{t-N_p+2} \\ 
\vdots \\ 
\vect{u}_t 
\end{bmatrix} \,, \label{eq:b} \\
\vect{y}_p &:= \vect{Y}_{t-N_p+1, N_p, 1} = \begin{bmatrix} 
\vect{y}_{t-N_p+1} \\ 
\vect{y}_{t-N_p+2} \\ 
\vdots \\ 
\vect{y}_t 
\end{bmatrix} \,. \label{eq:c}
\end{align}
\end{subequations}

Here, $\vect y$ is a vector that contains the indoor temperatures $T_{in}^1 \cdots T_{in}^4$ of four different rooms in the house. $\vect u$ is the vector containing outdoor temperature $T_{out}$ and power inputs to four the heat pumps $P^1 \cdots P^4$ respectively. In this example, we consider the scenario of heat pumps only being used for heating the house and not for cooling. $\vect f$ is single step linear data-driven predictor that at time-step $t$ takes $N_P$ step past inputs $u_P$ and past outputs $y_P$ to predict the next output $y_{t+1}$ at time-step $t+1$. $\epsilon \sim \mathcal{N}(\mu, \sigma^2)$ is an innovation term modeled as normal distribution based on the residuals of the data after fitting the linear model. Note that the ARX model \eqref{eq:house} will serve as the simulation environment and act as a proxy for the real system in this experiment.

\subsubsection{The predictive model}
Now we consider the deterministic part of the house system \eqref{eq:house} as a predictive digital of the system, 
\begin{equation}\label{eq:DT_house}
  \hat{\vect{y}}_{t+1} = \vect f \begin{bmatrix} \vect{u}_p \\ \vect{y}_p \end{bmatrix} \,.
\end{equation}
The only difference between the predictive model and its real system is the stochastic term $\epsilon$ estimated from the real-world data. We will show that by adapting the deterministic predictive model \eqref{eq:DT_house} according to \eqref{Th:necesary_condition} we can improve the performance of decision-making using the predictive model. The goal of the decision-making is to reduce the cost of electricity consumption for heating the house while providing a comfortable indoor temperature between 18 and 23 degrees. An \ac{mpc} scheme is formulated to solve this decision-making problem using the deterministic predictive model \eqref{eq:DT_house}.

\subsubsection{Decision-making problem formulation}
The model-based decision-making scheme is formulated as an \ac{mpc} scheme:
\begin{subequations}
\label{eq:house_MPC}
\begin{align}
\max_{\vect{\hat{y}}, \vect{u}} & \sum_{i=0}^{N-1} r \left(\vect{\hat{y}}_i, \vect{u}_i\right) + \gamma^{N} T\left(\vect{x}_N\right) \label{eq:MPC0:Cost}\,, \\
\mathrm{s.t.} & \quad \vect{\hat{y}}_{t+1} = \vect f \begin{bmatrix} \vect{u}_p \\ \vect{y}_p \end{bmatrix}\,, \\
&\quad 18 \leq \vect{\hat{y}}_{t+1} \leq 23 \label{eq:MPC0:Const}\,, \\
&\quad \vect{y}_{0} = \vect{y}_k \label{eq:MPC0:Boundaries}\,.
\end{align}
\end{subequations}
The reward $r$ is defined to minimize the cost of energy consumption based on the hourly market price $p$ and to maintain the indoor temperature around a set point $T_s = 21^\circ\mathrm{C}$:
\begin{equation}\label{eq:reward_smarthome}
  r (y, u) = - \alpha p  \vect P -  (\vect T_s - \vect T_i) \beta (\vect T_s - \vect T_i)^T\,.
\end{equation}
Here $\vect P$ represents the sum of power consumption from all four rooms and $\vect T$ is a vector representation of the temperature in all four rooms. The first term in \eqref{eq:reward_smarthome} $\alpha p  \vect P $, represents the total cost of electricity, and the second term $(\vect T_s - \vect T_i) \beta (\vect T_s - \vect T_i)^T$, represents the total comfort cost. The \ac{mpc} scheme \eqref{eq:house_MPC}  utilizes a best-fit, linear expected-value predictive model estimated through linear regression. Since the real system dynamics are linear, the linear predictive model can provide close to optimal decisions on controlling the heat pump. To apply the Theorem \eqref{Th:necesary_condition} directly to find the predictive model that provides optimal performance is computationally intractable due to the curse of dimensionality in solving the underlying optimal value function. Therefore, we apply the proposed practical approach of using \ac{rl} to modify the data-fitted linear predictive model \eqref{eq:DT_house} into a decision-oriented predictive model. We expect that such a modification will improve the closed-loop performance of the decision-making scheme \eqref{eq:house_MPC}.

\begin{figure}[t]
  \centering
  \begin{tikzpicture}
    \begin{axis}[
        width=\columnwidth,
        height=0.2\textwidth,
        xlabel={},
        ylabel={Price},
        ylabel style={font=\normalfont}, 
        xmin=0, xmax=30, 
        xtick={0,5,10,15,20,25,30}, 
        legend pos=north east, 
        grid=major,
        legend style={draw=none},
    ]
    
    \addplot[
        color=black, thick
    ] table [
        x expr=\coordindex, 
        y=mean_price, 
        col sep=comma
    ] {Figures/house/N1d_H6h/data.csv};
    
    \addplot [
        name path=upper, draw=none
    ] table [
        x expr=\coordindex, 
        y expr=\thisrow{mean_price} + sqrt(\thisrow{var_price}),
        col sep=comma
    ] {Figures/house/N1d_H6h/data.csv};

    \addplot [
        name path=lower, draw=none
    ] table [
        x expr=\coordindex, 
        y expr=\thisrow{mean_price} - sqrt(\thisrow{var_price}),
        col sep=comma
    ] {Figures/house/N1d_H6h/data.csv};

    \addplot[black!50, fill opacity=0.2] fill between[of=upper and lower];

    \end{axis}
\end{tikzpicture}

\vspace{0.3cm}
\begin{tikzpicture}
    \begin{axis}[
        width=\columnwidth,
        height=0.2\textwidth,
        xlabel={},
        ylabel={Comfort},
        ylabel style={font=\normalfont}, 
        xmin=0, xmax=30, 
        xtick={0,5,10,15,20,25,30}, 
        legend pos=north east, 
        grid=major,
        legend style={draw=none},
    ]
    
    \addplot[
        color=black, thick
    ] table [
        x expr=\coordindex, 
        y=mean_comfort, 
        col sep=comma
    ] {Figures/house/N1d_H6h/data.csv};
    
    \addplot [
        name path=upper, draw=none
    ] table [
        x expr=\coordindex, 
        y expr=\thisrow{mean_comfort} + sqrt(\thisrow{var_comfort}),
        col sep=comma
    ] {Figures/house/N1d_H6h/data.csv};

    \addplot [
        name path=lower, draw=none
    ] table [
        x expr=\coordindex, 
        y expr=\thisrow{mean_comfort} - sqrt(\thisrow{var_comfort}),
        col sep=comma
    ] {Figures/house/N1d_H6h/data.csv};

    \addplot[black!50, fill opacity=0.2] fill between[of=upper and lower];

    \end{axis}
\end{tikzpicture}

\vspace{0.3cm}
\begin{tikzpicture}
    \begin{axis}[
        width=\columnwidth,
        height=0.2\textwidth,
        xlabel={Learning iteration}, 
        ylabel={Total cost},
        ylabel style={font=\normalfont}, 
        xmin=0, xmax=30, 
        xtick={0,5,10,15,20,25,30}, 
        legend pos=north east, 
        grid=major,
        legend style={draw=none},
    ]
    
    \addplot[
        color=black, thick
    ] table [
        x expr=\coordindex, 
        y=mean_total, 
        col sep=comma
    ] {Figures/house/N1d_H6h/data.csv};
    
    \addplot [
        name path=upper, draw=none
    ] table [
        x expr=\coordindex, 
        y expr=\thisrow{mean_total} + sqrt(\thisrow{var_total}),
        col sep=comma
    ] {Figures/house/N1d_H6h/data.csv};

    \addplot [
        name path=lower, draw=none
    ] table [
        x expr=\coordindex, 
        y expr=\thisrow{mean_total} - sqrt(\thisrow{var_total}),
        col sep=comma
    ] {Figures/house/N1d_H6h/data.csv};

    \addplot[black!50, fill opacity=0.2] fill between[of=upper and lower];

    \end{axis}
\end{tikzpicture}
  \caption{Evaluation results for \ac{mpc} Scheme 1 with model adaptation. The plot illustrates the mean performance of 10 distinct \ac{mpc} schemes, each trained with a different random seed, evaluated after each learning iteration over 10 days of operation on a separate evaluation dataset.}
  \label{fig:house_6H}
\end{figure}
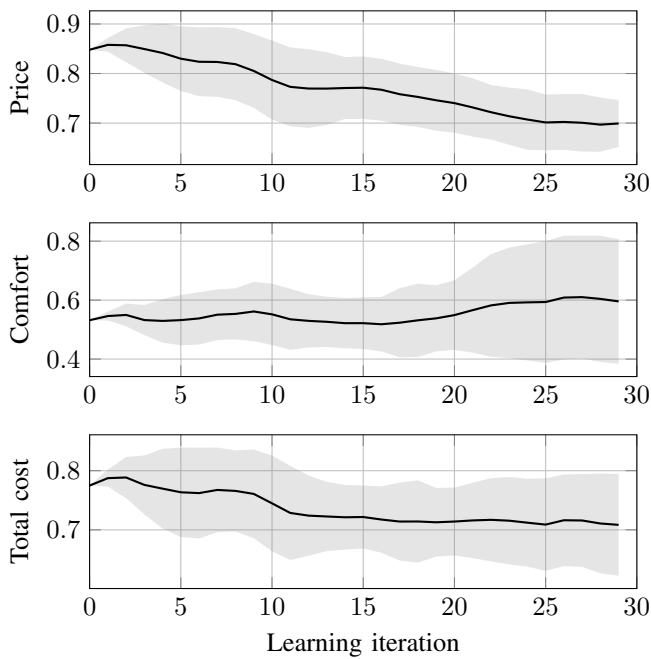

\subsubsection{RL-based fine-tuning of the expected-value model}
\ac{rl}-based \ac{mpc} approach is used to adapt the parameters of the linear deterministic predictive model in the \ac{mpc} scheme \eqref{eq:house_MPC} \cite{kordabad2023reinforcement}. The \ac{rl} algorithm used is \ac{lstdq} \cite{mahadevan2012representation} with linear critic. The \ac{mpc} horizon is 12 hours, the learning interval is 24 hours, and the learning duration is 90 days. The learning rate is \(e^{-10}\). Both \ac{mpc} sampling time and the predictive model sampling time are set to 5 minutes. However, to reduce computational effort, the \ac{mpc} scheme is solved only every 30 minutes, during which the first 6 actions are applied sequentially to the system instead of only the first action. The predictive model utilizes 1 hour of historical data to generate single-step future predictions. A total of 10 predictive models were estimated using 10 random seeds. Each seed randomly sampled the learning data from 8 months between 01-09-2017 and 30-04-2018, based on weather data for the Norwegian city of Trondheim and electricity spot market price data. The learning process consists of 30 iterations.

The resulting decision-oriented predictive models were evaluated for four random seeds, each representing a randomly chosen 10-day evaluation period.  During the evaluation, 10 \ac{mpc} schemes with 10 different decision-oriented predictive models from each of the 30 learning iterations were tested on the real system, and their performance in terms of comfort and price savings was monitored. The results of the 10 predictive models over 4 random evaluation seeds were averaged and presented in Fig.~\ref{fig:house_6H}. We observe that deriving the decision-oriented predictive model through fine-tuning of the data-fitted predictive model has led to a performance improvement of approximately 9\%.

\section{Discussion}\label{sec:discussion}
A key insight from the conditions for optimality is that constructing decision-oriented predictive models for decision-making requires looking beyond conventional data-fitting approaches. Predictive models for decision-making should be constructed based on Theorem~\ref{Th:necesary_condition}, with the data-fitted model as a support or prior. In practice, this translates to relying on decision-oriented approaches to build predictive models. For example, using \ac{rl} to fine-tune models estimated through \eqref{eq:Perf:SYSID} or any of the standard parameter estimation approaches discussed in Section \ref{sec:dt for sdm} could offer a viable solution.  This approach is demonstrated in the smart home example in Section \ref{sec:smarthome}, and can be extended to any general model-based decision-making scheme if we ensure the differentiability of the decision-making scheme. Interestingly, deterministic models can be an ideal choice for decision-oriented predictive models, as they can achieve optimality even for stochastic systems while being easier to handle in model-based optimization compared to probabilistic models. This observation is validated in the smart home energy management example in Section \ref{sec:smarthome}, where a deterministic decision-oriented predictive model could effectively account for the stochasticity of the real system. Given the local optimality of expected-value models for dissipative problems, we emphasize that constructing decision-oriented predictive models is particularly important for non-dissipative economic problems.

Regarding the properties of the decision-oriented model, even though the unicity can be achieved with the support of the conventional data fitting approach \eqref{eq:Perf:SYSID}, there is no guarantee regarding preserving or enforcing the properties of linearity, continuity, or smoothness on the resulting predictive model. For instance, the optimal predictive model for a linear system with stochasticity may turn out to be non-linear in nature as demonstrated in example \ref{sec:BatExample}.  Our empirical experiments show that a specific choice of parameters, for example, $\Delta$ in sufficient condition \eqref{eq:suficient_condition} ensures the continuity of the resulting predictive model. Further research is needed on identifying conditions to enforce such properties within the context of the Theorem~\ref{Th:necesary_condition}. Moreover, the optimality conditions may not be interesting for planning problems where capturing the open-loop planning reward is more important than satisfying the single-step criterion in \eqref{eq:necessary_condition}. Additionally, the presented optimality conditions on the predictive model do not extend readily to multi-step predictive models.

Although the optimality of the decision-making process discussed in this work is directly tied to closed-loop performance, we have not specifically addressed constraint satisfaction or safety aspects. However, according to Theorem~\ref{Th:necesary_condition}, decision-oriented predictive models capture worst-case scenarios for safe decision-making—an approach consistent with common practice. We have based our analysis purely on the \ac{mdp} framework, where decision-making is in the sense of the expected value of the outcomes. While in general most decision-making problems can be approached using the expected value approach it may not be suitable where the decision-making is inherently risky.  Risk-sensitive optimality criteria for \ac{mdp}s have been considered by various authors over the years and often lead to non-standard \ac{mdp}s which cannot be solved in a straightforward way by using the Bellman equation \cite{bauerle2011markov, rigter2022planning, ahmadi2023risk}. This needs further research in understanding optimality beyond standard \ac{mdp} under Bellman optimality conditions

\section{Conclusions}\label{sec:conclusion}
We presented the sufficient and necessary conditions for a predictive model to achieve optimality in decision-making within the \ac{mdp} framework. These conditions are counter-intuitive to the conventional notion of constructing predictive models based on prediction accuracy. We have established why it is important to look beyond prediction accuracy, especially when using predictive models for policy optimization problems with economic objectives. We refer to predictive models designed for optimal decision-making performance as decision-oriented predictive models. Our main message for constructing a decision-oriented predictive model is to align it with the performance measures of the decision-making problem while leveraging the support of conventional data-fitting approaches. However, constructing such predictive models directly from the necessary or sufficient optimality conditions can be complex or even impractical for high-dimensional systems. Instead, indirect approaches like \ac{rl}-based fine-tuning, incorporating the decision-making objective into the loss function, or combining offline \ac{rl}, with the support of conventional data-fitting approach may offer more practical solutions. A key direction for future work is to develop such algorithmic solutions for constructing decision-oriented predictive models to apply the theory to complex real-world problems.

\bibliographystyle{IEEEtran}
\bibliography{main} 

\appendices
\section{Proof of Lemma~\ref{lem:necessary_condition_A}} \label{apx:proof_necessary_lemma}
In the proof, we will consider the solution to \ac{mdp} as a cost minimization problem. The resulting condition is then translated into a reward maximization problem, as expressed in the necessary condition \eqref{eq:necesary_condition_A} in Lemma~\ref{lem:necessary_condition_A}. We do this simply by using disadvantage functions defined as the negative of the corresponding advantage functions: $D^\star(\vect s, \vect a) = - A^\star(\vect s, \vect a)$ and $\hat{D}^\star(\vect s, \vect a) = - \hat{A}^\star(\vect s, \vect a)$. The corresponding \eqref{eq:Advantage_equality} in terms of disadvantage function is given by:
\begin{equation}
    \min _{\mathbf{a}} \hat{D}^\star(\mathbf{s}, \mathbf{a})=\min _{\mathbf{a}} D^{\star}(\mathbf{s}, \mathbf{a})=0 \quad \forall \vect s  \,.
\end{equation}
The following two Lemmas help us to prove the Lemma~\ref{lem:necessary_condition_A}.
\begin{Lemma}\label{lem:necessary_condition_1}
  The condition: there exists a class K-function $\alpha\,:\,\mathbb R_+\mapsto \mathbb R_+$ such that
  \begin{equation}\label{eq:aplha_for_A}
     \hat{D}^\star(\mathbf{s}, \mathbf{a}) \geq \alpha (D^{\star}(\mathbf{s}, \mathbf{a})), \quad \forall\, \vect s\,, \vect a
  \end{equation}
  is necessary and sufficient condition for 
  \begin{equation}\label{eq:A_min_subset}
      \underset{\vect a}{\arg \min } \hat{D}^\star(\mathbf{s}, \mathbf{a}) \subseteq \underset{\vect a}{\arg \min } D^{\star}(\mathbf{s}, \mathbf{a}), \quad \forall \vect s
  \end{equation}
  to hold.
\end{Lemma}

\begin{IEEEproof}
  We will prove it by  establishing that (i) $\neg$ \eqref{eq:A_min_subset} $\Rightarrow$ $\neg$ \eqref{eq:aplha_for_A} and (ii) \eqref{eq:A_min_subset} $\Rightarrow$ \eqref{eq:aplha_for_A}. 
  
  Let us first establish $\neg$ \eqref{eq:A_min_subset} $\Rightarrow$ $\neg$ \eqref{eq:aplha_for_A} . $\neg$ \eqref{eq:A_min_subset}  implies that there exist a state-action pair $\overline{\vect s}, \overline{\vect a} $ such that 

  \begin{equation}
      \bar{\vect a}\in \argmin_{\vect a}  \hat{D}^\star\left(\bar{\vect s},\vect a\right) ,\quad\text{and}\quad \bar{\vect a}\notin \argmin_{\vect a} D^\star\left(\bar{\vect s},{\vect a}\right),
      \end{equation}
i.e.,        
  \begin{equation}
      \hat{D}^\star(\overline{\mathbf{s}}, \overline{\mathbf{a}})=0, \quad \text { and } \quad D^{\star}(\overline{\mathbf{s}}, \overline{\mathbf{a}})>0 \,.
  \end{equation}
  Then \eqref{eq:aplha_for_A} can not hold for any K-function $\alpha$ at $\overline{\vect s}, \overline{\vect a}$. 

  Next we establish that \eqref{eq:A_min_subset} $\Rightarrow$ \eqref{eq:aplha_for_A}. We start by constructing the following function $\alpha_0$: 

  \begin{subequations}
          \label{eq:alpha:K}
          \begin{align}
              \alpha_0(x) :=
              \min_{ {\vect s, \vect a}} \quad &\hat{D}^\star\left({\vect s}, {\vect a}\right)  \\
              \mathrm{s.t.}\,\,\quad \quad & D^\star\left({\vect s}, {\vect a}\right) \geq  x\,
          \end{align}
      \end{subequations}
   for all $x\leq \bar D^\star$, where 
   \begin{align}
      \bar D^\star=\max_{ {\vect s, \vect a}} D^\star (\vect s,\vect a)\,,
   \end{align}
  and:
   \begin{align}\label{eq:alpha:K:2}
       \alpha_0(x) := \alpha_0(\bar D^\star)+x-\bar D^\star, \quad \forall\, x>\bar D^\star\,.
   \end{align}
  We now prove the following properties for $\alpha_0$:
  \begin{enumerate}
     \item\label{item:alpha0:1} $\alpha_0(0)=0$.
      \item \label{item:alpha0:2} $\alpha_0(x)>0$ for $x>0$.
      \item \label{item:alpha0:3} $\alpha_0(x)$ is a (not necessarily strictly) increasing function.
      \item \label{item:alpha0:4} $\alpha_0(x)$ satisfies \eqref{eq:aplha_for_A}.
  \end{enumerate}
  For condition \ref{item:alpha0:1}), the feasible domain of~\eqref{eq:alpha:K} for $x=0$, contains all $\vect s, \vect a$, including $(\vect s,\hat{\pi}^\star (\vect s))$, therefore  $\alpha_0(0)=0$. 
  
  We prove condition \ref{item:alpha0:2}) using contradiction. First, we observe that the objective function is non-negative for all $ x\in [0, \bar D^\star]$. Suppose that $\alpha_0 (x)=0$ for some $x\in (0, \bar D^\star]$. It results in there exists $\bar{\vect s}$ and $\bar{\vect a}$ such that $\hat{D}^\star\left(\bar {\vect s}, \bar {\vect a}\right)=0$ while respecting constraint  $D^\star\left(\bar {\vect s}, \bar {\vect a}\right)\geq x>0$ which is in contradiction with \eqref{eq:A_min_subset}. Note that for all $x>\bar D^\star$, $\alpha_0(x)> 0$ by construction in~\eqref{eq:alpha:K:2}.
  
  To prove \ref{item:alpha0:3}), we first define the set of feasible solutions of~\eqref{eq:alpha:K} at a given $x$ by $\mathfrak C (x)$. One can observe that $\mathfrak C (x_2)\subseteq \mathfrak C (x_1)$ for any $x_1 \leq x_2$, making the optimal objective (or $\alpha_0(x)$) increasing when $x$ increases. For for all $x>\bar D^\star$, we have $\alpha_0(x)\geq \alpha_0(\bar D^\star)$, and $\alpha_0(x)$ is also increasing by construction in~\eqref{eq:alpha:K:2}.
  
  We then prove \ref{item:alpha0:4}), i.e., we show that, for all $\vect s_0\in S$ and $\vect a_0\in C(\vect s_0)$, we have:
   \begin{align}
   \label{eq:Algebraicy:K0}
   \hat D^\star\left(\vect s_0,\vect a_0\right) \geq \alpha\left( D^\star\left(\vect s_0,\vect a_0\right)\right).
  \end{align}
  Since $D^\star\left(\vect s_0,\vect a_0\right)\leq \bar D^\star$,  from~\eqref{eq:alpha:K}, we have:
  \begin{subequations}
          \label{eq:alpha:K0}
          \begin{align}
              \alpha_0(D^\star\left(\vect s_0,\vect a_0\right)) :=
              \min_{{\vect s}, {\vect a}} \quad &\hat D^\star\left({\vect s}, {\vect a}\right)  \\
              \mathrm{s.t.}\,\,\quad \quad & D^\star\left({\vect s}, {\vect a}\right) \geq  D^\star\left(\vect s_0,\vect a_0\right).
          \end{align}
      \end{subequations}
  Note that $\vect s_0, \vect a_0$ is a feasible feasible solution to~\eqref{eq:alpha:K0}, proving~\eqref{eq:Algebraicy:K0} from the optimality. Furthermore, one can verify that any function satisfying conditions~\ref{item:alpha0:1}), \ref{item:alpha0:2}) and \ref{item:alpha0:3}) can be lower bounded by a K-function, concluding the proof.
  \end{IEEEproof}

\begin{Lemma} \label{lem:necessary_condition_2} The condition: there exists a K-function $\beta\,:\,\mathbb R_+\mapsto \mathbb R_+$ such that
  \begin{align}
  \label{eq:Algebraicy:rev:K}
  \beta(D^\star\left(\vect s,\vect a\right)) \geq \hat{D}^\star\left(\vect s,\vect a\right) ,\quad \forall\, \vect s,\vect a
 \end{align}
 is necessary and sufficient for
 \begin{align}
 \label{eq:AMinSetMatchy:rev:K}
 \argmin_{\vect a} D^\star\left(\vect s,\vect a\right) \subseteq \argmin_{\vect a}\hat{D}^\star \left(\vect s,\vect a\right),\quad \forall\, \vect s
 \end{align}
 to hold.
  \end{Lemma}

 \begin{IEEEproof} 
 We observe that Lemma \ref{lem:necessary_condition_2} implies that the condition: there exists a K-function $\tilde\beta\,:\,\mathbb R_+\mapsto \mathbb R_+$ such that
  \begin{align}
  \label{eq:Algebraicy:K2}
  D^\star\left(\vect s,\vect a\right) \geq \tilde\beta\left( \hat{D}^\star \left(\vect s,\vect a\right)\right),\quad \forall\, \vect s,\vect a
 \end{align}
 is necessary and sufficient for
 \begin{align}
 \label{eq:AMinSetMatchy:K2}
 \argmin_{\vect a} D^\star\left(\vect s,\vect a\right) \subseteq \argmin_{\vect a} \hat{D}^\star \left(\vect s,\vect a\right),\quad \forall\, \vect s
 \end{align}
 to hold. Since $\tilde\beta$ is a K-function, then there is a K-function $\beta\,:\,\mathbb R\to\mathbb R$ satisfying 
 \begin{align}
 \beta\left(\tilde \beta\left(x\right)\right) = x \,.
\end{align}
 Then applying $\beta$ on \eqref{eq:Algebraicy:K2} implies that  \eqref{eq:Algebraicy:K2} is equivalent to \eqref{eq:Algebraicy:rev:K}. Now, the proof of Lemma~\ref{lem:necessary_condition_2} directly follows from the proof of Lemma~\ref{lem:necessary_condition_1}, utilizing the equivalence between conditions \eqref{eq:Algebraicy:K2}–\eqref{eq:AMinSetMatchy:K2} and \eqref{eq:aplha_for_A}–\eqref{eq:A_min_subset} in Lemma~\ref{lem:necessary_condition_1}.
 \end{IEEEproof}

The proof of Lemma~\ref{lem:necessary_condition_A} is derived by combining Lemmas~\ref{lem:necessary_condition_1} and \ref{lem:necessary_condition_2}. This involves flipping the inequalities \eqref{eq:aplha_for_A} and \eqref{eq:Algebraicy:rev:K} to transform the disadvantage functions to advantage functions, as expressed in \eqref{eq:necesary_condition_A}.

  \section{Proof of Theorem~\ref{Th:necesary_condition}} \label{apx:proof_theorem}

  Expanding \eqref{eq:necesary_condition_A}, 
  \begin{equation}
      \alpha({Q}^\star(\mathbf{s}, \mathbf{a})- {V}^\star(\mathbf{s})) \geq \hat{Q}^\star(\mathbf{s}, \mathbf{a})-\hat{V}^\star(\mathbf{s}) \geq \beta (Q^{\star}(\mathbf{s}, \mathbf{a})- V^{\star}(\mathbf{s}))\,.    
  \end{equation}
  
  Hence,
  \begin{equation}\label{eq:condition_without_V_star}
  \begin{aligned}
  \alpha (r & (\mathbf{s}, \mathbf{a})+ \gamma \mathbb{E}\left[V^{\star}\left(\mathbf{s}_{+}\right) \mid \mathbf{s}, \mathbf{a}\right]- V^{\star}(\mathbf{s})) \geq \\
  & r(\mathbf{s}, \mathbf{a})+ \Lambda(\mathbf{s}, \mathbf{a})+\gamma \mathbb{E}\left[\hat{V}^{\star}\left(\mathbf{\hat{s}}_{+}\right) \mid \mathbf{s}, \mathbf{a}\right]-\hat{V}^\star(\mathbf{s}) \geq \\
  & \hspace{1cm} \beta (r(\mathbf{s}, \mathbf{a})+\gamma \mathbb{E}\left[ V^{\star}\left(\mathbf{s}_{+}\right) \mid \mathbf{s}, \mathbf{a}\right]- V^{\star}(\mathbf{s}))
  \end{aligned}
  \end{equation}

  Since $\hat{A}^\star$ is independent of $\lambda$, so that $\lambda$ can be chosen arbitrarily without impacting necessary condition \eqref{eq:necesary_condition_A} nor the factor $\alpha$ or $\beta$. It follows that (i) choosing a specific $\lambda$ does not weaken the necessary condition to a sufficient one, and (ii) the value of $\lambda$ can always be selected such that
  \begin{equation}\label{eq:lambda_for_alpha}
      \hat{V}^\star = V^\star.
  \end{equation}
  Resulting in the necessary condition \eqref{eq:necessary_condition},
  \begin{equation}
      \begin{aligned}
      \alpha (r & (\mathbf{s}, \mathbf{a})+ \gamma \mathbb{E}\left[V^{\star}\left(\mathbf{s}_{+}\right) \mid \mathbf{s}, \mathbf{a}\right]- V^{\star}(\mathbf{s})) \geq \\
      & r(\mathbf{s}, \mathbf{a})+ \Lambda(\mathbf{s}, \mathbf{a})+\gamma \mathbb{E}\left[{V}^{\star}\left(\mathbf{\hat{s}}_{+}\right) \mid \mathbf{s}, \mathbf{a}\right]-{V}^\star(\mathbf{s}) \geq \\
      & \hspace{1cm} \beta (r(\mathbf{s}, \mathbf{a})+\gamma \mathbb{E}\left[ V^{\star}\left(\mathbf{s}_{+}\right) \mid \mathbf{s}, \mathbf{a}\right]- V^{\star}(\mathbf{s}))
      \end{aligned}
  \end{equation}

  for all $\vect s, \vect a$.

\section{Proof of Corollary~\ref{Th:sufficient_condition}}\label{apx:proof_sufficeint}
Consider $\hat{Q}^{\star}$ of the model-based \ac{mdp}: 
  \begin{equation}
  \hat{Q}^{\star} (\vect s, \vect a) = r(\vect s,\ \vect a) + \gamma \mathbb E_{\hat\rho}\left[V^\star\left(\vect{\hat s}_+\right)\,|\, \vect s,\vect a\, \right] \label{eq:V_pi_model}
  \end{equation}
  Using \eqref{eq:suficient_condition},  we get,
  \begin{equation}
      \hat{Q}^{\star} (\vect s, \vect a) = r(\vect s,\ \vect a)  + \gamma \mathbb E_{\rho}\left[V^\star\left(\vect s_+\right)\,|\, \vect s,\vect a\, \right] - \gamma \Delta\,.
  \end{equation}
  This can be rearranged as:
  \begin{align}\label{eq:sufficient_Q_proof}
      \hat{Q}^{\star} (\vect s, \vect a) + \gamma \Delta = {Q}^{\star} (\vect s, \vect a) \,.
  \end{align}
  
   Therefore, with $Q_0 = \gamma \Delta$, \eqref{eq:sufficient_Q_proof} corresponds to the sufficient condition on predictive model optimality \eqref{eq:DT:PerfectModel:PlusConstant}.

\section{Proof of Corollary~\ref{cor:deterministic}}\label{apx:proof:deterministic}
Consider a deterministic model of the form:
\begin{equation}
    \hat \rho (\hat {\vect s}_+ | \vect s,\vect a) = \delta(\hat {\vect s}_+-\vect f(\vect s, \vect a))\,,
\end{equation}
where \(\delta\) denotes the Dirac delta function.  Applying commutativity of the $\mathbb{E}$ operator in $V^\star$, we get:
\begin{equation}\label{eq:value_model_to_value_mdp}
  \mathbb{E}_{\hat{\rho}}\left[{V}^\star\left(\hat{\vect{s}}_+\right)\,|\, \vect s,\vect a   \right] = {V}^\star\left(\vect f\left(\vect s,\vect a\right)\right)\,.
\end{equation}
The proof follows from substituting \eqref{eq:value_model_to_value_mdp} in \eqref{eq:necessary_condition}.

\section{Proof of Proposition \ref{prop:existance}}\label{apx:proof:existance}
Let us define:
\begin{align}
S\left(\vect s,\vect a\right) = \left\{\vect s_+\quad \text{s.t}\quad \rho\left(\vect s_+|\vect s,\vect a\right)> 0\right\}
\end{align}
the support set of $\rho$. We then observe that
\begin{align}
\inf_{\vect s_+\in S\left(\vect s,\vect a\right) } V_\star\left(\vect s_+\right) \leq \mathbb E\left[V_\star\left(\vect s_+\right)|\vect s,\vect a\right] \leq \sup_{\vect s_+\in S\left(\vect s,\vect a\right) } V_\star\left(\vect s_+\right) 
\end{align}
Because $S\left(\vect s,\vect a\right)$ is by assumption fully connected and $V_\star$ is continuous, then there is a point $\vect s_+^0 \in S\left(\vect s,\vect a\right)$ such that
\begin{align}
V_\star\left(\vect s_+^0\right)=  \mathbb E\left[V_\star\left(\vect s_+\right)|\vect s,\vect a\right]
\end{align}
Hence we can assign $\vect f\left(\vect s,\vect a\right)=\vect s_+^0$, and satisfy \eqref{eq:Opt} and \eqref{eq:Likelihood}.

\end{document}